\documentclass{article}

 \usepackage[main, final]{neurips_2026}

\usepackage[utf8]{inputenc} 
\usepackage[T1]{fontenc}    
\usepackage{hyperref}       
\usepackage{url}            
\usepackage{booktabs}       
\usepackage{amsfonts}       
\usepackage{nicefrac}       
\usepackage{microtype}      
\usepackage{xcolor}         
\usepackage{amsmath}
\usepackage{cleveref}
\usepackage{multirow}
\usepackage{subcaption}
\usepackage{graphicx}
\usepackage[table]{xcolor}
\usepackage{wrapfig}
\usepackage{booktabs}
\usepackage{multirow}
\usepackage{caption}
\usepackage{placeins}
\title{Trust Region Continual Learning \\as an Implicit Meta-Learner}

\author{
    Zekun Wang\thanks{Equal contribution.} $\quad$ Anant Gupta\footnotemark[1] $\quad$ Christopher J. MacLellan \\
    Georgia Institute of Technology\\
    \texttt{\{zekun,agupta886,cmaclell\}@gatech.edu}
}

\begin{document}

\maketitle


\begin{abstract}
Continual learning aims to acquire tasks sequentially without catastrophic forgetting, yet standard strategies face a core tradeoff: regularization-based methods (e.g., EWC) can overconstrain updates when task optima are weakly overlapping, while replay-based methods can retain performance but drift due to imperfect replay.
We study a hybrid perspective: \emph{trust region continual learning} that combines generative replay with a Fisher-metric trust region constraint.
We show that, under local approximations, the resulting update admits a Model-Agnostic Meta-Learning (MAML) interpretation with a single implicit inner step: replay supplies an old-task gradient signal (query-like), while the Fisher-weighted penalty provides efficient offline curvature shaping (support-like).
This yields an emergent meta-learning property in continual learning: the model becomes an initialization that rapidly \emph{re-converges} to prior task optima after each task transition, without explicitly optimizing a bilevel objective that couples inner adaptation with outer meta-updates.
Empirically, on task-incremental diffusion image generation and continual diffusion-policy control, trust region continual learning achieves the best final performance and retention, and accelerates early-task recovery compared with EWC, replay, and continual meta-learning baselines.
\end{abstract}

\section{Introduction}

Learning is rarely stationary. 
Models face shifting data streams and evolving environments. 
Continual learning studies how to acquire new tasks sequentially while remaining robust on previously learned tasks without repeatedly retraining from scratch.
However, neural networks trained with standard gradient-based updates can suffer severe catastrophic forgetting, motivating methods that explicitly stabilize old knowledge while learning new skills.
Two families of methods dominate continual learning.
Regularization-based approaches preserve earlier tasks by penalizing changes to important parameters, often using Fisher-like quadratic constraints \citep{kirkpatrick2017overcoming,zenke2017synaptic,aljundi2018memory}.
Yet, they rely critically on both the existence of shared parameters across tasks and the fidelity of their curvature approximations, conditions that often fail under distribution shift or crude Fisher estimates.
Replay-based approaches retain performance by interleaving current training with stored or generated past samples \citep{rolnick2019experience,shin2017continual}, but are limited by memory policies and can drift when generations are imperfect and bootstrapped \citep{zverev2025dangers}.

A natural extension is to combine replay and regularization so that replay encourages parameter sharing while regularization constrains replay-induced drift.
Recent diffusion-based generative models \cite{ho2020ddpm} make this combination particularly effective.
Diffusion models can produce high-quality replay samples, and their gradient geometry can yield substantially improved Fisher approximations: \cite{wang2025rank1fisher} shows that diffusion models admit an approximately rank-1 empirical Fisher, enabling a cheap yet informative Fisher estimate and making elastic weight consolidation (EWC) \cite{kirkpatrick2017overcoming} a strong complement to replay.
Inspired by this line of work, we adopt a trust region view of EWC and replay synergy: EWC as a Fisher-metric trust region while replay supplies the gradient signal that drives parameter updates, analogous in spirit to trust region methods that limit steps under local Kullback–Leibler divergence \citep{schulman2015trpo}.

In this paper, we analyze the learning dynamics induced by such updates on diffusion models.
We show that, under local approximations, the resulting update takes the form of Model-Agnostic Meta-Learning (MAML) with one inner adaptation step \citep{finn2017maml}: replay supplies a query gradient on past data, while the EWC Fisher forms an efficient offline approximation to the second-order Hessian matrix on the past support set.
This view links continual learning and meta-learning at the level of optimization dynamics.
Even without an explicit bilevel procedure that couples inner adaptation with outer meta-updates \cite{finn2017maml}, the continual learning update can yield an initialization-like property that supports faster re-adaptation on prior tasks.
Unlike continual meta-learning methods that begin with an explicit bi-level objective and then adapt it to sequential tasks \citep{finn2019onlinemeta,riemer2019mer,javed2019mrcl,gupta2020lamamllookaheadmetalearning,pmlr-v202-wu23d}, we start from this continual learning objective and analyze what meta-learning structure it induces.
Empirically, we test whether trust region continual learning yields \emph{faster re-convergence} under sequential training in two regimes: (i) low-heterogeneity task-incremental diffusion image generation on ImageNet-500, and (ii) high-heterogeneity diffusion policy control on Continual-World-10.
We compare against continual-learning baselines (generative replay \citep{pmlr-v274-masip25a}, \textsc{EWC} with rank-1 Fisher \citep{kirkpatrick2017overcoming,wang2025rank1fisher}, and naive fine-tuning) and continual meta-learning baselines (follow the meta leader (FTML) \cite{finn2019onlinemeta}, variance reduced meta-continual learning (VR-MCL) \cite{wu2024meta}).
Across both domains, trust region continual learning re-converges on earlier tasks faster and improves retention, supporting our claim that Fisher-metric trust region constraints can induce emergent meta-learning behavior.

Our contributions are: (i) We frame the combination of EWC and replay as trust region continual learning, where replay supplies old-task gradients and the EWC pulls updates to be near old-task optima \citep{kirkpatrick2017overcoming,schulman2015trpo};
(ii) Under local approximations, we show the trust region continual learning takes a one-step MAML form, exposing an implicit meta-objective over past tasks \citep{finn2017maml};
(iii) On diffusion image generation and diffusion policy control, we benchmark against \textsc{EWC}-only, replay-only, and continual meta-learning \citep{finn2019onlinemeta,wu2024meta}, showing faster early re-convergence with competitive retention.
More broadly, we propose a robust, scalable, and efficient approach that bridges continual learning and meta-learning.

\section{Related Work}

\subsection{Continual Learning}
Continual learning studies training on a stream of tasks while maintaining performance on earlier tasks,
where catastrophic forgetting is a central challenge \cite{mccloskey1989catastrophic,french1999catastrophic}.
A major family of methods mitigates forgetting by constraining parameter drift, e.g., EWC via a Fisher-weighted quadratic penalty \cite{kirkpatrick2017overcoming} and synaptic intelligence via
online importance estimates accumulated over training trajectories \cite{zenke2017synaptic}. Another family relies on
replay, interleaving stored examples from prior tasks with current-task updates \cite{rebuffi2017icarl,BARARI2026101447}.
Hybrid combinations of replay and regularization \cite{heng2023selective,wang2025rank1fisher} are also consistent with consolidation and replay perspectives from cognitive science \cite{mcclelland1995cls}.

Generative replay replaces explicit storage of past data with a learned generator that synthesizes pseudo-samples
from previous tasks, reducing memory demands and enabling flexible replay schedules \cite{shin2017continual,vandeven2018gr}.
Diffusion models are particularly attractive generators due to their high-fidelity synthesis; diffusion-based
generative replay has been explored in class-incremental generation and dense prediction
\cite{pmlr-v202-gao23e,chen2023diffusepast,kim2024sddgr}. Since the generator itself is continually updated, stabilizing
the reverse denoising dynamics becomes important; generative distillation addresses this by distilling the reverse
chain across timesteps \cite{pmlr-v274-masip25a}. 
Building on this line, we study hybrid continual diffusion training that combines diffusion replay with stronger curvature approximations derived from diffusion gradients \cite{wang2025rank1fisher}.
Online Curvature-Aware Replay (OCAR) \citep{pmlr-v267-urettini25a} uses online Fisher preconditioning, while we combine replay with rank-1 Fisher constraints around past task optima and analyze the resulting connection to MAML.

\subsection{Meta-Learning and Continual Meta-Learning}
Meta-learning (``learning to learn'') aims to acquire inductive biases from a distribution of tasks so that a model can adapt rapidly to a new task using only a few examples \cite{finn2017maml,hospedales2020metalearning}.
A common formulation is bilevel optimization: for task $\mathcal{T}\sim p(\mathcal{T})$, an inner loop adapts parameters using task training data, and an outer loop updates meta-parameters to minimize post-adaptation loss. With one inner step, the objective is:
\begin{equation}
\begin{aligned}
\min_{\boldsymbol{\theta}}~\mathbb{E}_{\mathcal{T}\sim p(\mathcal{T})}\!\Big[\mathcal{L}_{\mathcal{T}}\big(\theta';\,\mathcal{D}^{\mathrm{te}}_{\mathcal{T}}\big)\Big]
\quad \text{s.t.}\quad
\theta' \;=\; \theta-\alpha \nabla_{\theta}\mathcal{L}_{\mathcal{T}}(\theta;\mathcal{D}^{\mathrm{tr}}_{\mathcal{T}}).
\end{aligned}
\end{equation}
where $\mathcal{D}^{\mathrm{tr}}_{\mathcal{T}}$ and $\mathcal{D}^{\mathrm{te}}_{\mathcal{T}}$ are the task's training and test splits, $\theta$ is the meta-initialization, and $\theta'$ is obtained by taking a training gradient step of size $\alpha$. The outer objective minimizes the expected test loss at $\theta'$ \cite{finn2017maml}.
Follow-up work improves scalability by avoiding higher-order derivatives \cite{nichol2018firstorder} or by learning the update rule itself \cite{li2017metasgd}.

Continual meta-learning extends this paradigm to settings where tasks arrive sequentially (often with distributional shift), requiring the meta-learner to accumulate reusable knowledge while remaining adaptable.
Most prior continual meta-learning methods start from an explicit meta-learning objective and adapt it to sequential task streams, often using replay, online meta-updates, or meta-regularization \cite{finn2019onlinemeta,riemer2019mer,javed2019mrcl,gupta2020lamamllookaheadmetalearning}.
\cite{wu2024meta} further bridges meta-learning and continual learning approaches through Hessian approximation.
Regularization-based continual learning typically uses fixed (often diagonal) curvature surrogates for past tasks, whereas meta-gradient updates implicitly maintain an online Hessian estimate whose variance can be reduced via improved memory sampling \cite{wu2024meta}.
In our analysis, EWC provides an efficient offline Fisher surrogate for this curvature, especially for diffusion models where the empirical Fisher closely tracks local Hessian structure \cite{wang2025rank1fisher}.
Unlike prior bi-level formulations, we start from a practical continual learning objective and analyze the meta-learning structure it already induces.

\section{Continual Learning in Trust Regions Induces Meta-Learning}
In this section, we first define our continual learning setup and where the meta-learning behavior could emerge in \cref{sec:setup}, then recast continual learning as learning within a trust region around task optima in \cref{sec:cl}, and finally reinterpret the resulting updates as a MAML-style optimization procedure in continual learning in \Cref{sec:cml}.

\subsection{Continual Learning Problem Set-Up}
\label{sec:setup}

In continual learning, the learner observes a stream of tasks
$\{\mathcal{T}_t\}_{t=1}^{T}$ sequentially.
Each task $\mathcal{T}_t$ comes with a training set $\mathcal{D}^{\mathrm{tr}}_t$
and a disjoint evaluation set $\mathcal{D}^{\mathrm{te}}_t$.
Let $f_\theta$ denote the model and let $\mathcal{L}_t(\theta;\mathcal{D})$ be the empirical loss on task $t$
evaluated on dataset $\mathcal{D}$.
The standard continual learning objective focuses on retention: after finishing task $t$, we want good performance on all tasks seen so far,
e.g., low $\frac{1}{t}\sum_{i=1}^{t}\mathcal{L}_i(\theta_t;\mathcal{D}^{\mathrm{te}}_i)$, while learning the new task.

Beyond final retained performance, we study a re-convergence phenomenon:
during the learning of new tasks, performance on an old task $\mathcal{T}_i$ may drop;
yet a strong continual learning solution should recover its earlier good performance quickly.
We say $\theta_t$ exhibits fast recovery on task $\mathcal{T}_i$ if the evaluation performance
returns near the task's previously achieved level using fewer update steps.

Fast recovery also means that $\theta_t$ functions as an initialization from which few gradient steps
rapidly (re-)attain good performance on old tasks.
This is the same structural property optimized in gradient-based meta-learning, which finds parameters that enable fast adaptation under a fixed inner-loop rule \cite{finn2017maml}.
Our setting studies an analogous fast adaptation problem, but under the continual learning constraint that tasks arrive sequentially and must not be forgotten.
The key difference is that traditional few-shot evaluation of meta-learning typically targets fast adaptation to new tasks drawn i.i.d.\ from a task distribution.
In contrast, our ``tasks of interest'' are the previously encountered tasks,
and $\theta_t$ must simultaneously acquire new tasks and preserve the ability to rapidly re-converge on old tasks.

\subsection{Continual Learning in Trust Regions}
\label{sec:cl}
We first revisit two canonical continual learning
strategies: EWC and generative replay through the lens of trust region optimization in
parameter space. 
Throughout, let $\theta_{i}^{\star}$ denote the end-of-task parameters after learning $\mathcal{T}_i$ in the continual run. For replay-based methods, this training uses the mixture of current-task and replay data.
Let $F^{(i)}_{\theta^*_i}$ be the Fisher information (evaluated at $\theta_i^\star$) used to quantify how sensitive
task $\mathcal{T}_i$ is to perturbations of $\theta$.

\paragraph{EWC as a quadratic trust region around past optima.}
From a Bayesian update perspective, EWC can be derived by applying a Laplace approximation to the previous posterior around the past optimum $\theta_i^\star$.
The resulting objective for learning
task $\mathcal{T}_t$ takes the form
$\min_{\theta}\;
\mathcal{L}_{\mathcal{T}_t}(\theta;\mathcal{D}_t)
\;+\;\frac{\lambda}{2}\sum_{i=1}^{t-1}
(\theta-\theta_i^{\star})^{\top}F^{(i)}_{\theta^*_i}(\theta-\theta_i^{\star})$,
where $\lambda>0$ trades off plasticity and stability. 
This is the Lagrangian
relaxation of an explicit trust region constraint:
\begin{equation}
\label{eq:trust-region}
\min_{\theta}\; \mathcal{L}_{\mathcal{T}_t}(\theta;\mathcal{D}_t)
\quad \text{s.t.}\quad
\sum_{i=1}^{t-1}(\theta-\theta_i^{\star})^{\top}F^{(i)}_{\theta^*_i}(\theta-\theta_i^{\star}) \le \delta,
\end{equation}
for some radius $\delta>0$. 
Intuitively, EWC pulls updates toward an ellipsoidal neighborhood around previous task optima, penalizing movement against the
directions that are deemed important by the Fisher.

\begin{wrapfigure}[30]{r}{0.48\textwidth}
    \vspace{-0.5\baselineskip}
    \centering

    \begin{subfigure}[t]{\linewidth}
        \centering
        \includegraphics[width=0.95\linewidth]{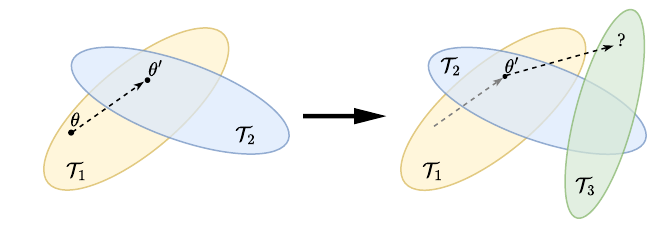}
        \caption{Elastic Weight Consolidation-based approach}
        \label{fig:ewc}
    \end{subfigure}

    \begin{subfigure}[t]{\linewidth}
        \centering
        \includegraphics[width=1.0\linewidth]{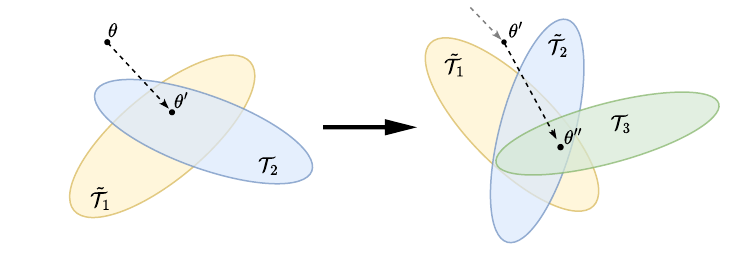}
        \caption{Generative replay-based approach}
        \label{fig:replay}
    \end{subfigure}

    \begin{subfigure}[t]{\linewidth}
        \centering
        \includegraphics[width=1.0\linewidth]{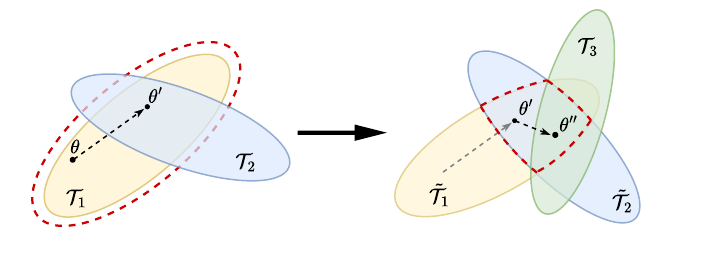}
        \caption{Trust region continual learning}
        \label{fig:trcl}
    \end{subfigure}

    \caption{Illustrations of parameter updates when learning a new task $\mathcal{T}_3$ under different continual-learning strategies. Colored ellipses denote low-loss regions for each task. Dark arrows show the current update direction from $\theta$, gray arrows indicate prior trajectory, and the red dashed region indicates the trust region constraint.}
    \label{fig:token-tsne-vertical}
    \vspace{-1.0\baselineskip}
\end{wrapfigure}
In practice, forming the full $F^{(i)}_{\theta^*_i}$ is often infeasible, and EWC typically uses a diagonal
approximation $F^{(i)}_{\theta^*_i}\approx \mathrm{diag}(F^{(i)}_{\theta^*_i})$. This simplification can be especially limiting
in diffusion models where curvature may concentrate in off-diagonal structure, making diagonal
constraints weak or even uninformative \cite{wang2025rank1fisher}.
A second limitation is geometric: if the new task optimum
lies outside (or far from) the intersection of these trust regions, then the feasible set in
\Cref{eq:trust-region} can be effectively empty, yielding the failure mode illustrated in~\Cref{fig:ewc}.

\paragraph{Generative replay as optimizing on a union of task data.}
Generative replay-based continual learning instead trains on a mixture of current-task data and replayed samples
from past tasks. 
Abstractly, let $\tilde{\mathcal{D}}_{k}$ denote replay data approximating task
$\mathcal{T}_k$. A generic replay objective is:
\begin{equation}
\label{eq:replay-generic}
\min_{\theta}\;
\mathcal{L}_{\mathcal{T}_t}(\theta;\mathcal{D}_t)
\;+\;\beta\sum_{i=1}^{t-1}\mathcal{L}_{\mathcal{T}_i}(\theta;\tilde{\mathcal{D}}_{i}),
\end{equation}
with $\beta\ge 0$ controlling the replay ratio.

Replay addresses EWC's ``disjoint optima'' issue by explicitly encouraging solutions that perform well
on the \emph{union} of (current + replayed) data.
This effectively explores shared low-dimensional data manifolds supported by all task datasets, often enabling convergence to a basin that overlaps
across tasks (\Cref{fig:replay}).
However, replay can drift due to imperfect generators or
compounding approximation error: even if the model continues to fit replayed samples, small replay
mismatches can accumulate and move parameters toward a different-but-valid basin that is far from
earlier optima in overparameterized networks, requiring more updates to converge to new optima.
\paragraph{Trust region replay and the role of a stronger Fisher.}
One can combine the complementary strengths of replay and EWC by enforcing trust region
updates while replay promotes cross-task overlap:
\begin{equation}
\label{eq:ewc-replay-general}
\min_{\theta}\;
\mathcal{L}_{\mathcal{T}_t}(\theta;\mathcal{D}_t)
\;+\;\beta\,\mathcal{L}_{\mathrm{Replay}}(\theta)
\;+\;\lambda\mathcal{L}_{\mathrm{EWC}}.
\end{equation}
This objective matches the geometry in \Cref{fig:trcl}: replay pulls optimization toward regions
that remain good for previous tasks (making the trust region non-vacuous), while the Fisher-weighted
penalty anchors the updates to be around past task optima.

The effectiveness of \Cref{eq:ewc-replay-general} depends critically on the quality of $F^{(i)}_{\theta^*_i}$.
Following \cite{wang2025rank1fisher}, diffusion models admit a cheap but informative Fisher
approximation: in the later diffusion timesteps, per-sample gradients become strongly collinear with their mean,
so the empirical Fisher is effectively rank-1. 
As a result, replay makes a shared basin available, while the rank-1 trust region makes movement within that basin stable by explicitly
constraining the dominant curvature direction. 
This trust region view will be the key bridge in
\Cref{sec:cml}, where we reinterpret the resulting continual updates as a MAML-style optimization.

\subsection{Reinterpretation as a MAML-style meta-learning}
\label{sec:cml}

We now ask \emph{what} our continual objective in \Cref{eq:ewc-replay-general} is implicitly learning.
Taking one gradient step on \Cref{eq:ewc-replay-general} yields:
\begin{equation}
\begin{aligned}
\label{eq:cml-update-decomp}
\theta \;\leftarrow\; \theta &\;-\;\eta\Big(
\underbrace{\nabla_{\theta}\mathcal{L}_{\mathcal{T}_t}(\theta;\mathcal{D}_t)}_{\textbf{(A) current-task fit}}\;+\;
\sum_{i=1}^{t-1}\underbrace{\beta\nabla_{\theta}\mathcal{L}_{\mathcal{T}_i}(\theta;\tilde{\mathcal{D}}_{i})}_{\textbf{(B) old-task replay}}\;+\;
\sum_{i=1}^{t-1}\underbrace{\lambda F^{(i)}_{\theta^*_i}(\theta-\theta^*_i)}_{\textbf{(C) old-task EWC}}
\Big),
\end{aligned}
\end{equation}
where $\eta$ is the step size, $\tilde{\mathcal{D}}_i$ denotes generative replayed data for old task $\mathcal{T}_i$,
and $F^{(i)}_{\theta^*_i}$ is the empirical Fisher estimated from task $\mathcal{T}_i$.

For standard MAML with a single inner step ($k=1$), the adapted parameters for task $\mathcal{T}_i$ are
\(
\theta'_i = \theta - \alpha \nabla_{\theta}\mathcal{L}_{\mathcal{T}_i}(\theta;\mathcal{D}^{\mathrm{tr}}_{\mathcal{T}_i}).
\)
The outer update minimizes the post-adaptation (query) loss:
\begin{equation}
\label{eq:maml-one-step-expanded}
\theta \leftarrow \theta \;-\;\eta\sum_{i=1}^{t-1}
\Big(I-\alpha H^{\mathrm{tr}}_{\theta_i}\Big)\,
\nabla_{\theta'}\mathcal{L}_{\mathcal{T}_i}\!\left(\theta'_i;\mathcal{D}^{\mathrm{te}}_{\mathcal{T}_i}\right)\nonumber,
\end{equation}
where $H^{\mathrm{tr}}_{\theta_i}=\nabla_{\theta}^2\mathcal{L}_{\mathcal{T}_i}(\theta;\mathcal{D}^{\mathrm{tr}}_{\mathcal{T}_i})$.
Expanding the product:
\begin{equation}
\begin{aligned}
\label{eq:maml-one-step-two-terms}
\theta \leftarrow \theta&\;-\;\eta\Big(
\sum_{i=1}^{t-1}\underbrace{\nabla_{\theta'}\mathcal{L}_{\mathcal{T}_i}\!\left(\theta'_i;\mathcal{D}^{\mathrm{te}}_{\mathcal{T}_i}\right)}_{\textbf{(I) \emph{query} gradient at adapted params}}\;-\;
\sum_{i=1}^{t-1}\underbrace{\alpha H^{\mathrm{tr}}_{\theta_i}\,
\nabla_{\theta'}\mathcal{L}_{\mathcal{T}_i}\!\left(\theta'_i;\mathcal{D}^{\mathrm{te}}_{\mathcal{T}_i}\right)}_{\textbf{(II) \emph{support} curvature correction}}
\Big).
\end{aligned}
\end{equation}

We compare the \textbf{old-task} part of the trust region update directly with the MAML meta-gradient. 
The connection to MAML relies on a trust region-style observation: for many old tasks $i<t$, the continual update keeps
$\theta$ in a low-loss neighborhood around $\theta_i^*$ (the low-error region in \Cref{fig:trcl}), so evaluating old-task gradients at
$\theta'_i$ versus at $\theta$ makes little difference to first order \cite{ghorbani2019investigation}.
Moreover, replayed samples provide a practical proxy for held-out performance \emph{query/test} on old tasks.
With these, evaluating the old-task query gradient at $\theta'_i$ is well-approximated by evaluating it at $\theta$:
\begin{equation}
\label{eq:first-conn}
\nabla_{\theta}\mathcal{L}_{\mathcal{T}_i}(\theta;\tilde{\mathcal{D}}_i)
\;
\approx
\nabla_{\theta'}\mathcal{L}_{\mathcal{T}_i}(\theta'_i;\mathcal{D}^{\mathrm{te}}_{\mathcal{T}_i})
\;=
\textbf{(I)},
\end{equation}
Define the old-task displacement
\(
\delta_i := \theta-\theta_i^*.
\)
A local Taylor expansion of the old-task query gradient around $\theta_i^*$ gives:
\begin{equation}
\label{eq:grad-local-linear}
\nabla_\theta \mathcal{L}_{\mathcal{T}_i}(\theta;\mathcal{D}^{\mathrm{te}}_{\mathcal{T}_i})
\;\approx\;
H^{\mathrm{te}}_{\theta^*_i}\,\delta_i.
\end{equation}
For negative log-likelihood objectives, the Fisher information equals the expected Hessian under the model distribution. Let $\ell(\theta;x)=-\log p_\theta(x)$ and $g(\theta;x)=\nabla_\theta\ell(\theta;x)$. Taking expectations over $x\sim p_\theta$ gives:
\begin{equation}
\label{eq:fisher-hessian-identity-main}
F_{\theta}
=
\mathbb{E}_{x}\!\left[g(\theta;x)g(\theta;x)^\top\right]
=
\mathbb{E}_{x}\!\left[\nabla_\theta^2 \ell(\theta;x)\right]
= \mathbb{E}_{x}\left[H_{\theta}\right],
\end{equation}
with a full derivation in Appendix~\ref{app:appendix-fisher-hessian}. This identity motivates using Fisher curvature for diffusion training \cite{sohl2015noneq,ho2020ddpm}. In practice, MAML computes $H^{\mathrm{tr}}_{\theta_i}$ from support-set mini-batches.
Under locality, curvature varies slowly within the neighborhood, so $H^{\mathrm{tr}}_{\theta_i}\approx H^{\mathrm{tr}}_{\theta_i^*}$, consistent with the observed stability of dominant Hessian directions near converged optima \cite{ghorbani2019investigation}.
Moreover, we assume $H^{\mathrm{tr}}_{\theta^*_i}\approx H^{\mathrm{te}}_{\theta^*_i}$ following the insight that the Fisher/Hessian characterizes the intrinsic geometric properties of the model’s learned predictive distribution \cite{martens2020naturalgradient}. 
Consequently, the local curvature remains similar across any samples (replayed or real) that align with this underlying manifold.
Therefore, by \Cref{eq:fisher-hessian-identity-main}, both can be approximated by a common Fisher curvature proxy evaluated at the old optimum:
\begin{equation}
\label{eq:curvature-identifications}
H^{\mathrm{tr}}_{\theta_i}
\;\approx\;
H^{\mathrm{tr}}_{\theta^*_i}
\;\approx\;
H^{\mathrm{te}}_{\theta^*_i}
\;\approx\;
F^{(i)}_{\theta^*_i}.
\end{equation}
Substituting \Cref{eq:grad-local-linear,eq:curvature-identifications} into \Cref{eq:cml-update-decomp}, the combined replay+EWC old-task update becomes:
\begin{equation}
\label{eq:bc-fisher-direction}
\textbf{(B)}+\textbf{(C)}
\;\approx\;
\beta F^{(i)}_{\theta_i^*}\delta_i + \lambda F^{(i)}_{\theta_i^*}\delta_i
\;=\;
(\beta+\lambda)F^{(i)}_{\theta_i^*}\delta_i.
\end{equation}
We now compare this with the MAML meta-gradient.
Using the same local query-gradient and curvature approximations from \Cref{eq:grad-local-linear,eq:curvature-identifications} gives:
\begin{equation}
\textbf{(I)}-\textbf{(II)}
\;\approx\;
F^{(i)}_{\theta^*_i}\delta_i-\alpha\left(F^{(i)}_{\theta^*_i}\right)^2\delta_i
\;=\;
\left(F^{(i)}_{\theta^*_i}
-
\alpha\left(F^{(i)}_{\theta^*_i}\right)^2\right)\delta_i.
\end{equation}


When $F^{(i)}_{\theta^*_i}$ is approximately rank-1 in diffusion models \cite{wang2025rank1fisher},
\(F^{(i)}_{\theta^*_i}\approx \rho_i u_i u_i^\top\) with \( \|u_i\|=1\), so
\(
\left(F^{(i)}_{\theta^*_i}\right)^2\approx \rho_iF^{(i)}_{\theta^*_i};
\)
see Appendix~\ref{app:appendix-fsquare-rank1}. Therefore, the MAML old-task meta-gradient under the rank-1 approximation is:
\begin{equation}
\label{eq:II-to-F2-main}
\textbf{(I)}-\textbf{(II)}
\;\approx\;
(1-\alpha\rho_i)F^{(i)}_{\theta^*_i}\delta_i.
\end{equation}

Equations~\eqref{eq:bc-fisher-direction} and \eqref{eq:II-to-F2-main} show that the combined replay+EWC old-task update and the MAML meta-gradient share the same Fisher-shaped direction, up to task-dependent scalar factors. The directions agree when the one-step inner update is locally stable along the dominant Fisher direction, \(1-\alpha\rho_i>0\), and more generally whenever the scalar multiplying \(F^{(i)}_{\theta_i^*}\delta_i\) is positive. Under the rank-1 form, \(\rho_i\) is also the Fisher scale, since \(\rho_i=\|F^{(i)}_{\theta_i^*}\|_F\). Thus, a task-dependent EWC weight can be interpreted as \(\lambda_i\propto\rho_i\), making the EWC contribution match the MAML descent direction up to scale.
In practice, we use a benchmark-wise constant \(\lambda\) set by the observed Fisher/gradient scale, since task Fisher scales are comparable within each benchmark. 

Finally, under local approximations, $\textbf{(B)}+\textbf{(C)}$ can be read as a MAML-style old-task meta-gradient $\textbf{(I)}-\textbf{(II)}$.
Substituting this Fisher-shaped old-task meta-gradient approximation into \Cref{eq:cml-update-decomp}, the trust-region replay update can be read as:
\[
\theta \;\leftarrow\; \theta \;-\; \eta\Big(
\nabla_\theta\mathcal{L}_{\mathcal{T}_t}(\theta;\mathcal{D}_t)
+
\nabla_\theta\mathcal{L}_{\mathrm{MAML}}(\theta_{i<t})
\Big).
\]
We refer to Appendix~\ref{app:empirical-analysis} for empirical analysis of these local approximations.

\section{Experimental Evaluations}
\label{sec:experiments}
We evaluate whether our continual learning approach yields \emph{faster convergence} when training sequentially across tasks in two settings with different heterogeneity: (i) low-heterogeneity task-incremental diffusion image generation on ImageNet-500, and (ii) high-heterogeneity continual diffusion-policy control on Continual-World-10.

\paragraph{Baselines.}
We compare our method against three continual-learning baselines:
(i) \textbf{Generative replay}, implemented via generative distillation, which has been shown effective for diffusion models \cite{pmlr-v274-masip25a};
(ii) \textbf{EWC} \cite{kirkpatrick2017overcoming} using the rank-1 approximation \cite{wang2025rank1fisher};
and (iii) \textbf{Na\"ive continual fine-tuning}, which sequentially trains on each task.
To benchmark against continual meta-learning, we include \textbf{FTML}, an online extension of MAML \cite{finn2019onlinemeta}, and \textbf{VR-MCL}~\cite{wu2024meta}.
For scalability, we implement these meta-learning baselines with the standard \emph{first-order} approximation, which typically matches second-order performance while avoiding the prohibitive cost of higher-order differentiation at our scale \cite{nichol2018firstorder}, as our model and batch sizes are substantially larger than those in prior VR-MCL experiment setups; we provide a detailed size/computation comparison and implementation notes in the Appendix~\ref{appdx:compute}.
Importantly, VR-MCL's control-variate update still reduces variance of the (first-order) stochastic meta-gradient.
For meta-learning baselines, we use generative replay to keep the memory footprint and data-access assumptions consistent with our continual learning setting.

\subsection{Task-Incremental Image Generation Set-up}
\label{sec:exp_image}

\paragraph{Datasets and Metrics.}
We use a 500-class ImageNet subset as a scaled up Tiny-ImageNet while keeping the compute footprint low~\citep{wu2017tinyimagenet,russakovsky2015imagenet,deng2009imagenet}.
We follow a task-incremental protocol by splitting the 500 classes into 10 disjoint tasks of 50 classes each.
We evaluate with Fr\'echet Inception Distance (FID)~\citep{heusel2017ttur} on a held-out split, reporting average FID across tasks and forgetting, defined as the change in a task's FID from when it is first learned to the end of training.
We refer to Appendix~\ref{appdx:imagenet} for additional details.

\paragraph{Implementation Details.}
We train conditional diffusion models with a standard denoising diffusion objective~\citep{ho2020ddpm,nichol2021improved} that condition on class labels.
For meta-learning baselines, we split each training batch evenly into a support/query pair.
We perform a single inner adaptation step so that the total number of parameter updates and the total number of training samples processed match our continual-learning setting.
We refer to Appendix~\ref{appdx:imagenet} for model details and training configurations.

\subsection{Continual Robotic Manipulation Set-up}
\label{sec:exp_control}
\paragraph{Datasets and Metrics.}
We evaluate continual robotic manipulation on Continual-World-10 (CW10)~\citep{wolczyk2021continualworld}, a 10-task sequence from Meta-World~\citep{yu2020metaworld} (e.g., \textit{push-wall}, \textit{close-window}). 
Unlike ImageNet, CW10 spans distinct manipulation skills with different dynamics and reward conditions, and thus exhibits substantially higher task heterogeneity.
Following standard practice in Continual-World, we report the success rate for each task, defined as the fraction of evaluation rollouts that satisfy the environment's success condition.
After each task, we evaluate on all tasks seen so far using 100 trajectories per task and report average success.
Forgetting is the drop in a task's success rate from when it is first learned to the end of training.
See Appendix~\ref{appdx:cw10} for CW10 details.

\paragraph{Implementation Details.}
We instantiate the control policy as a Diffusion Policy that generates action chunks conditioned on recent observations~\citep{chi2023diffusionpolicy}.
To collect expert demonstrations, we roll out scripted experts provided by the benchmark for 2500 trajectories per task, yielding an offline trajectory dataset for each task.
Each training sample uses an observation horizon of 6 steps and predicts an action chunk of 2 steps, forming a sequence length of 8; we selected these values via a grid search over horizon/chunk combinations, with full results reported in Appendix~\ref{appdx:cw10}.
For meta-learning baselines, we similarly split each training batch evenly into a support/query pair and perform a single inner adaptation step to match our continual-learning setting.
We refer to Appendix~\ref{appdx:cw10} for model details and training configurations.

\subsection{Results and Discussions}
\paragraph{Trust region continual learning reduces catastrophic forgetting.}
We study whether Trust Region combines the strengths of EWC and replay by comparing it to its components and finetuning across two datasets with different task heterogeneity.
High task heterogeneity, such as CW10, contains meaningfully diverse manipulation tasks with varying degrees of transfer and interference, and prior analyses explicitly use transfer matrices to quantify when tasks reuse versus overwrite features \citep{wolczyk2021continualworld}.
More broadly, feature reuse and forgetting depend on task similarity: dissimilar tasks tend to transfer less and interfere more, making local constraints around old optima harder to satisfy \citep{YosinskiCBL14,lee2021impact,goldfarb2024tasksim}.

\begin{wraptable}[18]{r}{0.54\textwidth}
    \vspace{-0.8\baselineskip}
    \centering
    \scriptsize
    \setlength{\tabcolsep}{4pt}
    \renewcommand{\arraystretch}{1.1}
    \captionsetup{skip=4pt}

    \caption{Final performance and forgetting on ImageNet-500 and CW10 with standard errors. For ImageNet, we report average FID across tasks ($\mathcal{A}\mathrm{FID}$, lower is better) and average forgetting $\mathcal{F}$ (lower is better). For CW10, we report average Success rate (SR, higher is better) and average forgetting $\mathcal{F}$ (lower is better).}
    \label{tab:imagenet_fid_forgetting}

    \begin{tabular}{lcccc}
    \toprule
    \multirow{2}{*}{\textbf{Methods}} &
    \multicolumn{2}{c}{\textbf{ImageNet-500}} &
    \multicolumn{2}{c}{\textbf{CW10}} \\
    \cmidrule(lr){2-3}\cmidrule(lr){4-5}
    & $\mathbf{\mathcal{A}\mathrm{FID}}\downarrow$
    & $\mathbf{\mathcal{F}}\downarrow$
    & $\mathbf{\mathrm{SR}}\uparrow$
    & $\mathbf{\mathcal{F}}\downarrow$ \\
    \midrule
    Finetuning   & 86.2$_{\pm 7.3}$   & 50.4$_{\pm 7.2}$   & 17.9\%$_{\pm 2.4}$ & 73.0$_{\pm 2.0}$ \\
    EWC          & 77.2$_{\pm 1.2}$   & 41.2$_{\pm 2.3}$   & 17.6\%$_{\pm 1.8}$ & 73.7$_{\pm 1.3}$ \\
    Replay       & 53.4$_{\pm 6.0}$   & 18.2$_{\pm 4.6}$   & 85.3\%$_{\pm 2.0}$ & 8.2$_{\pm 2.0}$ \\
    \midrule
    FTML         & 172.5$_{\pm 9.1}$  & 128.5$_{\pm 9.0}$  & 78.5\%$_{\pm 3.8}$ & 13.3$_{\pm 3.3}$ \\
    VRMCL        & 142.2$_{\pm 8.5}$  & 96.6$_{\pm 8.4}$   & 77.9\%$_{\pm 1.5}$ & 11.7$_{\pm 2.0}$ \\
    \midrule
    Trust Region & \textbf{44.5$_{\pm 2.3}$} & \textbf{10.6$_{\pm 3.0}$} & \textbf{88.3\%$_{\pm 0.4}$} & \textbf{4.4$_{\pm 0.9}$} \\
    \bottomrule
    \end{tabular}

    \vspace{-1.0\baselineskip}
\end{wraptable}

\Cref{tab:imagenet_fid_forgetting} shows that Trust Region achieves the best final performance and the lowest forgetting on both datasets.
On high-heterogeneity CW10, Trust Region attains the highest average success rate ({88.3\%$_{\pm 0.4}$}) while also exhibiting the smallest magnitude of forgetting ({4.4$_{\pm 0.9}$}), improving over replay (85.3\%$_{\pm 2.0}$, $8.2_{\pm 2.0}$) and outperforming regularization-only baselines. 
EWC performs similarly to finetuning on CW10, consistent with the view that a local quadratic surrogate can be a weak guide when tasks have limited overlap \citep{kirkpatrick2017overcoming}.
Replay is competitive on CW10 because mixing past and current samples approximates multitask training over the union of tasks and thus directly supplies gradients for retention.
However, generative replay can suffer from error accumulation and distribution drift when earlier generators are imperfect, which compounds over the task sequence and can worsen forgetting \citep{pmlr-v202-gao23e,pmlr-v274-masip25a}.
Trust Region mitigates this by anchoring updates within a neighborhood that preserves past-task performance, improving stability relative to replay alone. 

On low-heterogeneity ImageNet-500, Trust Region outperforms replay, achieving the best average FID ({44.5$_{\pm 2.3}$}) and the lowest forgetting ({10.6$_{\pm 3.0}$}), improving over replay (53.4$_{\pm 6.0}$, 18.2$_{\pm 4.6}$) and outperforming finetuning and EWC.
This suggests that, even when tasks share more structure, the additional local constraint effectively helps prevent “mode hopping” across many equivalent low-loss solutions during long-horizon training in overparameterized generative models \citep{garipov2018modeconnectivity,draxler2018nobarriers}.

\begin{figure*}[t]
    \centering
    \begin{subfigure}[t]{1.0\linewidth}
        \centering
        \includegraphics[width=\linewidth]{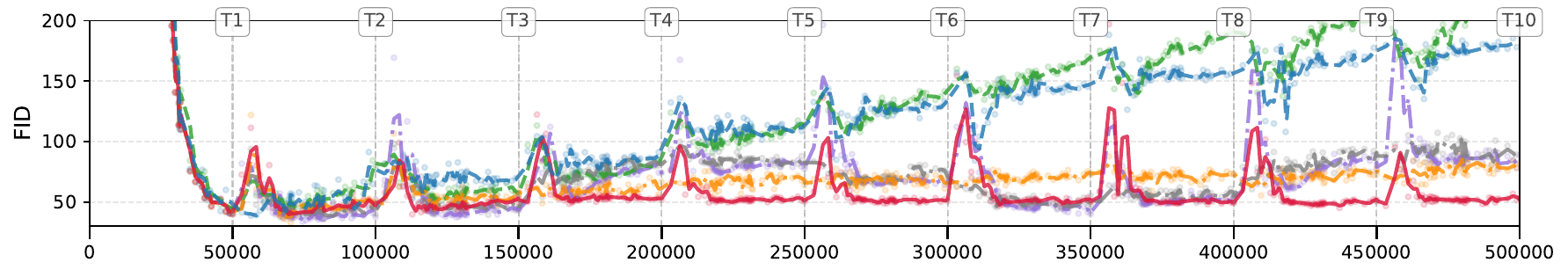}
        \caption{FID (lower is better) of task 1 while continual learning other tasks on ImageNet-500. Averaged over 3 random seeds.}
        \label{fig:imagenet-task1}
    \end{subfigure}
    \begin{subfigure}[t]{1.0\linewidth}
        \centering
        \includegraphics[width=\linewidth]{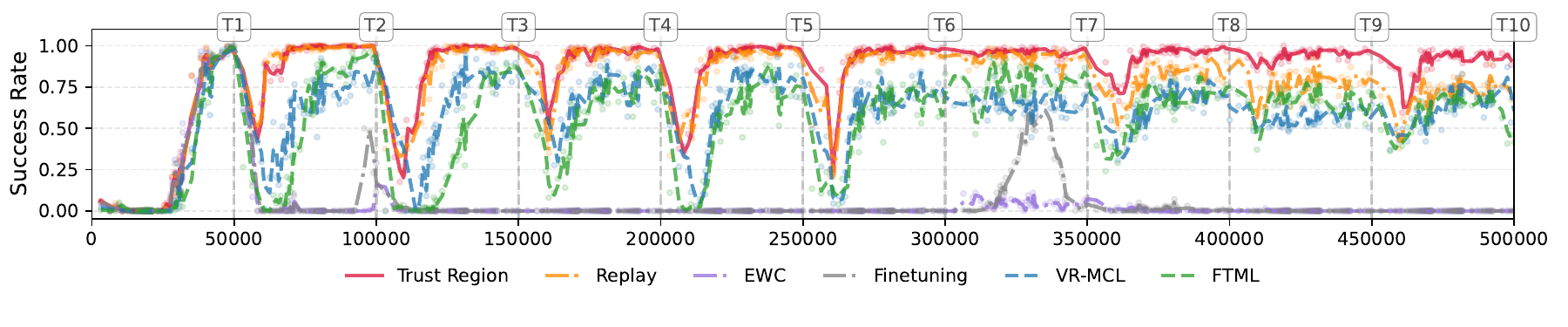}
        \caption{Success rate (higher is better) of task 1 while continual learning other tasks on CW10. Averaged over 4 random seeds.}
        \label{fig:cw10-task1}
    \end{subfigure}
    \caption{Task 1 performance over the course of continual training on 10 tasks, averaged over random seeds, on two datasets. X-axis: gradient update steps.}
    \label{fig:task1}
\end{figure*}

Surprisingly, continual meta-learning baselines underperform on ImageNet-500, falling well below finetuning: FTML reaches 172.5$_{\pm 9.1}$ average $\mathrm{FID}$ with 128.5$_{\pm 9.0}$ forgetting, and VRMCL reaches 142.2$_{\pm 8.5}$ with 96.6$_{\pm 8.4}$ forgetting.
Yet on CW10 their performance remains good, though still below Trust Region.
Intuitively, lower task heterogeneity implies greater overlap among task optima, so replay-only training already tracks a shared low-loss region.
Bilevel meta-learning then becomes more error-amplifying where imperfections in replay bias the outer meta-gradient objective and simultaneously corrupt the inner adaptation step taken from that slightly off initialization, pushing updates away from the shared basin (see \Cref{fig:imagenet-task1}).
In CW10, optima overlap less and inner adaptation is less informative, so replay supplies most of the retention signal.
Performance is therefore dominated by the outer meta-gradient objective, making the adaptation step comparatively less sensitive.
We refer to Appendices~\ref{app:hyperparameter-analysis} and~\ref{app-subsec:permutation} for hyperparameter and task-order experiments.

\paragraph{Trust region continual learning recovers early-task performance faster.}

\definecolor{ImBlue}{RGB}{110,180,255}
\definecolor{CwOrange}{RGB}{255,200,130}

\newcommand{\IMa}{ImBlue!15} \newcommand{\IMb}{ImBlue!25} \newcommand{\IMc}{ImBlue!35}
\newcommand{\IMd}{ImBlue!45} \newcommand{\IMe}{ImBlue!55} \newcommand{\IMf}{ImBlue!65}
\newcommand{\IMg}{ImBlue!75} \newcommand{\IMh}{ImBlue!85} \newcommand{\IMi}{ImBlue!95}

\newcommand{\CWa}{CwOrange!15} \newcommand{\CWb}{CwOrange!25} \newcommand{\CWc}{CwOrange!35}
\newcommand{\CWd}{CwOrange!45} \newcommand{\CWe}{CwOrange!55} \newcommand{\CWf}{CwOrange!65}
\newcommand{\CWg}{CwOrange!75} \newcommand{\CWh}{CwOrange!85} \newcommand{\CWi}{CwOrange!95}

\newcommand{\cc}[2]{\cellcolor{#1}\strut #2}

\begin{table}[t]
\centering
\tiny
\setlength{\tabcolsep}{1.9pt}
\renewcommand{\arraystretch}{1.15}
\caption{\textbf{Steps to re-converge Task~1 under continual learning.}
Each entry reports the number of gradient updates for Task~1 performance to return to the target degradation threshold after each task transition.
Smaller is better; ``--'' indicates that the target was not reached during training on that task.}
\begin{tabular}{l*{27}{c}}
\toprule

 & \multicolumn{27}{c}{\textbf{ImageNet-500}} \\
\cmidrule(lr){2-28}
\textbf{Method}
& \multicolumn{9}{c}{$+10\%$} & \multicolumn{9}{c}{$+20\%$} & \multicolumn{9}{c}{$+30\%$} \\
\cmidrule(lr){2-10}\cmidrule(lr){11-19}\cmidrule(lr){20-28}
& \cc{\IMa}{T2}  & \cc{\IMb}{T3}  & \cc{\IMc}{T4}  & \cc{\IMd}{T5}  & \cc{\IMe}{T6}  & \cc{\IMf}{T7}  & \cc{\IMg}{T8}  & \cc{\IMh}{T9}  & \cc{\IMi}{T10}
& \cc{\IMa}{T2}  & \cc{\IMb}{T3}  & \cc{\IMc}{T4}  & \cc{\IMd}{T5}  & \cc{\IMe}{T6}  & \cc{\IMf}{T7}  & \cc{\IMg}{T8}  & \cc{\IMh}{T9}  & \cc{\IMi}{T10}
& \cc{\IMa}{T2}  & \cc{\IMb}{T3}  & \cc{\IMc}{T4}  & \cc{\IMd}{T5}  & \cc{\IMe}{T6}  & \cc{\IMf}{T7}  & \cc{\IMg}{T8}  & \cc{\IMh}{T9}  & \cc{\IMi}{T10} \\
\midrule

Finetune
& \cc{\IMa}{25} & \cc{\IMb}{45} & \cc{\IMc}{--} & \cc{\IMd}{--} & \cc{\IMe}{--} & \cc{\IMf}{--} & \cc{\IMg}{--} & \cc{\IMh}{--} & \cc{\IMi}{--}
& \cc{\IMa}{5}  & \cc{\IMb}{35} & \cc{\IMc}{--} & \cc{\IMd}{--} & \cc{\IMe}{--} & \cc{\IMf}{35} & \cc{\IMg}{50} & \cc{\IMh}{--} & \cc{\IMi}{--}
& \cc{\IMa}{2}  & \cc{\IMb}{25} & \cc{\IMc}{--} & \cc{\IMd}{--} & \cc{\IMe}{--} & \cc{\IMf}{25} & \cc{\IMg}{50} & \cc{\IMh}{--} & \cc{\IMi}{--} \\

EWC
& \cc{\IMa}{20} & \cc{\IMb}{50} & \cc{\IMc}{--} & \cc{\IMd}{--} & \cc{\IMe}{--} & \cc{\IMf}{650} & \cc{\IMg}{--} & \cc{\IMh}{--} & \cc{\IMi}{--}
& \cc{\IMa}{15} & \cc{\IMb}{50} & \cc{\IMc}{--} & \cc{\IMd}{--} & \cc{\IMe}{--} & \cc{\IMf}{85}  & \cc{\IMg}{55} & \cc{\IMh}{--} & \cc{\IMi}{--}
& \cc{\IMa}{8}  & \cc{\IMb}{30} & \cc{\IMc}{55} & \cc{\IMd}{--} & \cc{\IMe}{--} & \cc{\IMf}{55}  & \cc{\IMg}{45} & \cc{\IMh}{--} & \cc{\IMi}{--} \\

Replay
& \cc{\IMa}{30} & \cc{\IMb}{90} & \cc{\IMc}{--} & \cc{\IMd}{--} & \cc{\IMe}{--} & \cc{\IMf}{--} & \cc{\IMg}{--} & \cc{\IMh}{--} & \cc{\IMi}{--}
& \cc{\IMa}{30} & \cc{\IMb}{75} & \cc{\IMc}{--} & \cc{\IMd}{--} & \cc{\IMe}{--} & \cc{\IMf}{--} & \cc{\IMg}{--} & \cc{\IMh}{--} & \cc{\IMi}{--}
& \cc{\IMa}{2}  & \cc{\IMb}{2}  & \cc{\IMc}{5}  & \cc{\IMd}{--} & \cc{\IMe}{--} & \cc{\IMf}{--} & \cc{\IMg}{--} & \cc{\IMh}{--} & \cc{\IMi}{--} \\

\midrule

FTML
& \cc{\IMa}{45} & \cc{\IMb}{--} & \cc{\IMc}{--} & \cc{\IMd}{--} & \cc{\IMe}{--} & \cc{\IMf}{--} & \cc{\IMg}{--} & \cc{\IMh}{--} & \cc{\IMi}{--}
& \cc{\IMa}{10} & \cc{\IMb}{80} & \cc{\IMc}{--} & \cc{\IMd}{--} & \cc{\IMe}{--} & \cc{\IMf}{--} & \cc{\IMg}{--} & \cc{\IMh}{--} & \cc{\IMi}{--}
& \cc{\IMa}{5}  & \cc{\IMb}{55} & \cc{\IMc}{--} & \cc{\IMd}{--} & \cc{\IMe}{--} & \cc{\IMf}{--} & \cc{\IMg}{--} & \cc{\IMh}{--} & \cc{\IMi}{--} \\

VRMCL
& \cc{\IMa}{4}  & \cc{\IMb}{--} & \cc{\IMc}{--} & \cc{\IMd}{--} & \cc{\IMe}{--} & \cc{\IMf}{--} & \cc{\IMg}{--} & \cc{\IMh}{--} & \cc{\IMi}{--}
& \cc{\IMa}{4}  & \cc{\IMb}{--} & \cc{\IMc}{--} & \cc{\IMd}{--} & \cc{\IMe}{--} & \cc{\IMf}{--} & \cc{\IMg}{--} & \cc{\IMh}{--} & \cc{\IMi}{--}
& \cc{\IMa}{4}  & \cc{\IMb}{15} & \cc{\IMc}{--} & \cc{\IMd}{--} & \cc{\IMe}{--} & \cc{\IMf}{--} & \cc{\IMg}{--} & \cc{\IMh}{--} & \cc{\IMi}{--} \\

\midrule

Trust Region
& \cc{\IMa}{35} & \cc{\IMb}{15} & \cc{\IMc}{--} & \cc{\IMd}{--} & \cc{\IMe}{--} & \cc{\IMf}{--} & \cc{\IMg}{--} & \cc{\IMh}{--}  & \cc{\IMi}{--}
& \cc{\IMa}{20} & \cc{\IMb}{15} & \cc{\IMc}{--} & \cc{\IMd}{--} & \cc{\IMe}{--} & \cc{\IMf}{50} & \cc{\IMg}{--} & \cc{\IMh}{350} & \cc{\IMi}{2}
& \cc{\IMa}{20} & \cc{\IMb}{15} & \cc{\IMc}{2}  & \cc{\IMd}{35} & \cc{\IMe}{35} & \cc{\IMf}{25} & \cc{\IMg}{55} & \cc{\IMh}{40}  & \cc{\IMi}{2} \\

\midrule\midrule

 & \multicolumn{27}{c}{\textbf{CW10}} \\
\cmidrule(lr){2-28}
\textbf{Method}
& \multicolumn{9}{c}{$99\%$} & \multicolumn{9}{c}{$90\%$} & \multicolumn{9}{c}{$80\%$} \\
\cmidrule(lr){2-10}\cmidrule(lr){11-19}\cmidrule(lr){20-28}
& \cc{\CWa}{T2}  & \cc{\CWb}{T3}  & \cc{\CWc}{T4}  & \cc{\CWd}{T5}  & \cc{\CWe}{T6}  & \cc{\CWf}{T7}  & \cc{\CWg}{T8}  & \cc{\CWh}{T9}  & \cc{\CWi}{T10}
& \cc{\CWa}{T2}  & \cc{\CWb}{T3}  & \cc{\CWc}{T4}  & \cc{\CWd}{T5}  & \cc{\CWe}{T6}  & \cc{\CWf}{T7}  & \cc{\CWg}{T8}  & \cc{\CWh}{T9}  & \cc{\CWi}{T10}
& \cc{\CWa}{T2}  & \cc{\CWb}{T3}  & \cc{\CWc}{T4}  & \cc{\CWd}{T5}  & \cc{\CWe}{T6}  & \cc{\CWf}{T7}  & \cc{\CWg}{T8}  & \cc{\CWh}{T9}  & \cc{\CWi}{T10} \\
\midrule

Finetune
& \cc{\CWa}{--} & \cc{\CWb}{--} & \cc{\CWc}{--} & \cc{\CWd}{--} & \cc{\CWe}{--} & \cc{\CWf}{--} & \cc{\CWg}{--} & \cc{\CWh}{--} & \cc{\CWi}{--}
& \cc{\CWa}{--} & \cc{\CWb}{--} & \cc{\CWc}{--} & \cc{\CWd}{--} & \cc{\CWe}{--} & \cc{\CWf}{--} & \cc{\CWg}{--} & \cc{\CWh}{--} & \cc{\CWi}{--}
& \cc{\CWa}{--} & \cc{\CWb}{--} & \cc{\CWc}{--} & \cc{\CWd}{--} & \cc{\CWe}{--} & \cc{\CWf}{--} & \cc{\CWg}{--} & \cc{\CWh}{--} & \cc{\CWi}{--} \\

EWC
& \cc{\CWa}{--} & \cc{\CWb}{--} & \cc{\CWc}{--} & \cc{\CWd}{--} & \cc{\CWe}{--} & \cc{\CWf}{--} & \cc{\CWg}{--} & \cc{\CWh}{--} & \cc{\CWi}{--}
& \cc{\CWa}{--} & \cc{\CWb}{--} & \cc{\CWc}{--} & \cc{\CWd}{--} & \cc{\CWe}{--} & \cc{\CWf}{--} & \cc{\CWg}{--} & \cc{\CWh}{--} & \cc{\CWi}{--}
& \cc{\CWa}{--} & \cc{\CWb}{--} & \cc{\CWc}{--} & \cc{\CWd}{--} & \cc{\CWe}{--} & \cc{\CWf}{--} & \cc{\CWg}{--} & \cc{\CWh}{--} & \cc{\CWi}{--} \\

Replay
& \cc{\CWa}{60} & \cc{\CWb}{95} & \cc{\CWc}{30} & \cc{\CWd}{80} & \cc{\CWe}{60} & \cc{\CWf}{30} & \cc{\CWg}{10000} & \cc{\CWh}{--} & \cc{\CWi}{--}
& \cc{\CWa}{10} & \cc{\CWb}{50} & \cc{\CWc}{2}  & \cc{\CWd}{20} & \cc{\CWe}{30} & \cc{\CWf}{2}  & \cc{\CWg}{35}    & \cc{\CWh}{2}  & \cc{\CWi}{8000}
& \cc{\CWa}{10} & \cc{\CWb}{35} & \cc{\CWc}{2}  & \cc{\CWd}{20} & \cc{\CWe}{15} & \cc{\CWf}{2}  & \cc{\CWg}{4}     & \cc{\CWh}{2}  & \cc{\CWi}{45} \\

\midrule

FTML
& \cc{\CWa}{--} & \cc{\CWb}{--} & \cc{\CWc}{--} & \cc{\CWd}{--} & \cc{\CWe}{--} & \cc{\CWf}{--} & \cc{\CWg}{--} & \cc{\CWh}{--} & \cc{\CWi}{--}
& \cc{\CWa}{550} & \cc{\CWb}{20000} & \cc{\CWc}{--} & \cc{\CWd}{3000} & \cc{\CWe}{--} & \cc{\CWf}{30} & \cc{\CWg}{--} & \cc{\CWh}{--} & \cc{\CWi}{--}
& \cc{\CWa}{250} & \cc{\CWb}{3000}  & \cc{\CWc}{700} & \cc{\CWd}{800}  & \cc{\CWe}{300} & \cc{\CWf}{2}  & \cc{\CWg}{150} & \cc{\CWh}{900} & \cc{\CWi}{--} \\

VRMCL
& \cc{\CWa}{--} & \cc{\CWb}{850} & \cc{\CWc}{10000} & \cc{\CWd}{--} & \cc{\CWe}{--} & \cc{\CWf}{--} & \cc{\CWg}{--} & \cc{\CWh}{--} & \cc{\CWi}{--}
& \cc{\CWa}{750} & \cc{\CWb}{650} & \cc{\CWc}{500}   & \cc{\CWd}{250} & \cc{\CWe}{--} & \cc{\CWf}{--} & \cc{\CWg}{--}  & \cc{\CWh}{--} & \cc{\CWi}{--}
& \cc{\CWa}{400} & \cc{\CWb}{350} & \cc{\CWc}{100}   & \cc{\CWd}{250} & \cc{\CWe}{85} & \cc{\CWf}{40} & \cc{\CWg}{550} & \cc{\CWh}{--} & \cc{\CWi}{30000} \\

\midrule

Trust Region
& \cc{\CWa}{60} & \cc{\CWb}{80} & \cc{\CWc}{45} & \cc{\CWd}{70} & \cc{\CWe}{55} & \cc{\CWf}{50} & \cc{\CWg}{2000} & \cc{\CWh}{--} & \cc{\CWi}{100}
& \cc{\CWa}{10} & \cc{\CWb}{45} & \cc{\CWc}{2}  & \cc{\CWd}{30} & \cc{\CWe}{30} & \cc{\CWf}{2}  & \cc{\CWg}{25}   & \cc{\CWh}{2}  & \cc{\CWi}{2}
& \cc{\CWa}{10} & \cc{\CWb}{35} & \cc{\CWc}{2}  & \cc{\CWd}{15} & \cc{\CWe}{4}  & \cc{\CWf}{2}  & \cc{\CWg}{4}    & \cc{\CWh}{2}  & \cc{\CWi}{2} \\
\bottomrule \\

\end{tabular}

\label{tab:steps}

\end{table}

To quantify how quickly early-task performance returns after learning new tasks, we measure re-convergence using the step-to-threshold metric in \Cref{tab:steps}: after each task transition in \Cref{fig:task1}, we count the number of gradient updates until Task~1 returns to a target level relative to its \emph{initial} optimum.
For ImageNet-500, \(\mathrm{FID}_1 \le (1+\tau)\mathrm{FID}_1^{(0)}\) with \(\tau\in\{10\%,20\%,30\%\}\), and for CW10, \(\mathrm{SR}_1 \ge \alpha\,\mathrm{SR}_1^{(0)}\) with \(\alpha\in\{99\%,90\%,80\%\}\).
Across tasks, these step counts typically range from single-digit to tens of updates for Trust Region methods (e.g., often recovers within \(\le 55\) steps on ImageNet and \(\le 45\) steps for CW10 at the \(90\%/80\%\) thresholds), while unstable baselines can exhibit orders-of-magnitude longer recovery tails (up to \(10^4\) steps) or fail to re-converge (``--'').

This behavior is informative about \emph{where} updates live in parameter space.
Replay provides gradients that keep old-task behavior ``reachable,'' but without an explicit locality constraint it can drift among multiple solutions that fit the replay mixture yet are less aligned with the original Task~1 basin, requiring more steps to re-attain the previous Task~1 optimum.
Trust Region sharpens this by constraining each update to remain in a neighborhood that preserves low error on past tasks, so the model does not need to ``re-learn'' Task~1 after each transition---it only needs a small correction to re-enter the original low-loss region.

Table~\ref{tab:steps} also helps explain an otherwise surprising pattern on ImageNet-500: finetuning can occasionally look competitive, and sometimes even stronger than EWC or Replay, at intermediate task transitions.
This does not indicate better forgetting robustness.
Rather, it reflects incidental positive transfer when the current task happens to be semantically or visually similar to Task~1.
In our default ImageNet-500 order, Task~1 is dominated by animal categories, and Tasks~2--3 are likewise animal-centric; later, Tasks~7--8 again contain animal-related classes.
In such cases, naive finetuning can partially preserve or recover Task~1 performance through shared visual structure, even without explicitly protecting old knowledge.
However, this advantage is fragile: once later tasks become less related to Task~1, finetuning quickly deteriorates and fails to re-converge under stricter thresholds.
Accordingly, these temporary gains should be interpreted as accidental cross-task similarity rather than genuine robustness to forgetting.

The contrast is even clearer on CW10.
In \Cref{fig:cw10-task1}, finetuning collapses to near-zero Task~1 success on later tasks, and EWC provides little improvement, consistent with stronger task heterogeneity where local regularization alone cannot maintain competence.
We do observe occasional transfer in finetuning (e.g., a transient improvement around task~6), suggesting some shared sub-skills, but this effect is sporadic and does not prevent eventual collapse.
In comparison, FTML and VR-MCL retain good Task~1 performance on CW10 but recover more slowly than Trust Region after each transition, indicating that an explicit ``fast-adaptation'' objective does not automatically translate to fast \emph{re-convergence} under long-horizon continual updates when replay and initialization drift accumulate.

\section{Conclusion}
We presented a trust region view of hybrid continual learning that unifies generative replay and EWC: replay keeps past-task behavior reachable by providing direct training signal on previous tasks, while the Fisher-metric penalty constrains each update to remain within a neighborhood that preserves low error on earlier optima.
Our analysis shows that this simple continual learning objective induces a MAML-style optimization form under local approximations, explaining why it can exhibit meta-learning-like behavior (fast re-adaptation) without an explicit bilevel procedure.
Across low-heterogeneity ImageNet-500 and high-heterogeneity CW10, trust region continual learning improves retention and accelerates old-task recovery, achieving the strongest final performance and the least forgetting.
More broadly, our results suggest a scalable continual and meta-learning paradigm for large models and demanding tasks like robotic control.

\paragraph{Limitations} One limitation is our implicit assumption that the task 1 optimum lies close to a shared optimum that also works well for later tasks (see \cref{fig:trcl}). 
When tasks are highly heterogeneous, this shared optimum may be far from the task-1 optimum, weakening the benefits of our approach.
We refer to Appendices~\ref{app:empirical-analysis} and~\ref{app:cw10-sensitivity} for additional analysis of replay fidelity, Fisher estimation, and the rank-1 approximation.
Future work could study how to identify or construct candidate tasks that are more likely to share a common parameter region for future tasks.


\bibliographystyle{abbrv}
\bibliography{main}


\appendix

\section{Full derivation of \Cref{eq:fisher-hessian-identity-main}.}
\label{app:appendix-fisher-hessian}
We now derive the identity linking Fisher and Hessian using the same notation.
Let $p_\theta(x)$ be the model distribution and define the per-example negative log-likelihood
\(
\ell(\theta;x) := -\log p_\theta(x).
\)
Then
\[
g(\theta;x)=\nabla_\theta \ell(\theta;x)= -\nabla_\theta \log p_\theta(x).
\]
Consider the Hessian:
\begin{align}
\nabla_\theta^2 \ell(\theta;x)
&= -\nabla_\theta^2 \log p_\theta(x)\nonumber\\
&= -\nabla_\theta\!\left(\frac{\nabla_\theta p_\theta(x)}{p_\theta(x)}\right)\nonumber\\
&= -\left(\frac{\nabla_\theta^2 p_\theta(x)}{p_\theta(x)} - \frac{\nabla_\theta p_\theta(x)\nabla_\theta p_\theta(x)^\top}{p_\theta(x)^2}\right)\nonumber\\
&= \frac{\nabla_\theta p_\theta(x)\nabla_\theta p_\theta(x)^\top}{p_\theta(x)^2}
\;-\;\frac{\nabla_\theta^2 p_\theta(x)}{p_\theta(x)}\nonumber\\
&= \big(\nabla_\theta \log p_\theta(x)\big)\big(\nabla_\theta \log p_\theta(x)\big)^\top
\;-\;\frac{\nabla_\theta^2 p_\theta(x)}{p_\theta(x)}.
\label{eq:hess-decomp}
\end{align}
Taking expectation over $x\sim p_\theta$,
\begin{align}
\mathbb{E}_{x\sim p_\theta}\!\left[\nabla_\theta^2 \ell(\theta;x)\right]
&=
\mathbb{E}_{x\sim p_\theta}\!\left[\big(\nabla_\theta \log p_\theta(x)\big)\big(\nabla_\theta \log p_\theta(x)\big)^\top\right]
-
\mathbb{E}_{x\sim p_\theta}\!\left[\frac{\nabla_\theta^2 p_\theta(x)}{p_\theta(x)}\right]\nonumber\\
&=
\mathbb{E}_{x\sim p_\theta}\!\left[\big(\nabla_\theta \log p_\theta(x)\big)\big(\nabla_\theta \log p_\theta(x)\big)^\top\right]
-
\int \nabla_\theta^2 p_\theta(x)\,dx
\label{eq:exp-step}
\end{align}
where we used $\mathbb{E}_{x\sim p_\theta}\!\left[\frac{\nabla_\theta^2 p_\theta(x)}{p_\theta(x)}\right]
=\int p_\theta(x)\frac{\nabla_\theta^2 p_\theta(x)}{p_\theta(x)}dx=\int \nabla_\theta^2 p_\theta(x)\,dx$.
Finally, because $\int p_\theta(x)\,dx=1$,
\begin{equation}
\int \nabla_\theta^2 p_\theta(x)\,dx
=
\nabla_\theta^2 \int p_\theta(x)\,dx
=
\nabla_\theta^2 1
=
0.
\label{eq:norm-constant}
\end{equation}
Substituting \Cref{eq:norm-constant} into \Cref{eq:exp-step} and using $g(\theta;x)=-\nabla_\theta\log p_\theta(x)$ yields
\[
\mathbb{E}_{x\sim p_\theta}\!\left[\nabla_\theta^2 \ell(\theta;x)\right]
=
\mathbb{E}_{x\sim p_\theta}\!\left[g(\theta;x)g(\theta;x)^\top\right],
\]
which is exactly \Cref{eq:fisher-hessian-identity-main}.
\section{Full derivation of \Cref{eq:II-to-F2-main}}
\label{app:appendix-fsquare-rank1}

Let $\delta_i=\theta-\theta_i^*$. If $\theta$ remains in a local neighborhood of $\theta_i^*$ (trust-region),
then a first-order Taylor expansion gives
\begin{equation}
\nabla_\theta \mathcal{L}_{\mathcal{T}_i}(\theta;\tilde{\mathcal{D}}_i)
=
\nabla_\theta \mathcal{L}_{\mathcal{T}_i}(\theta_i^*;\tilde{\mathcal{D}}_i)
+
H^{\mathrm{te}}_i(\theta_i^*)\,\delta_i
+
\mathcal{O}(\|\delta_i\|^2).
\end{equation}
Assuming $\theta_i^*$ is an (approximate) optimum for task $\mathcal{T}_i$, the first term is near zero, yielding
$\nabla_\theta \mathcal{L}_{\mathcal{T}_i}(\theta;\tilde{\mathcal{D}}_i)\approx H^{\mathrm{te}}_i(\theta_i^*)\delta_i$.

The MAML correction term is $H^{\mathrm{tr}}_i(\theta)\nabla_{\theta'}\mathcal{L}_{\mathcal{T}_i}(\theta';\mathcal{D}^{\mathrm{te}}_{\mathcal{T}_i})$.
Under the trust-region assumption, $H^{\mathrm{tr}}_i(\theta)\approx H^{\mathrm{tr}}_i(\theta_i^*)$.
Under replay, $\tilde{\mathcal{D}}_i$ is generated to match the old-task data distribution, so the query curvature
$H^{\mathrm{te}}_i(\theta_i^*)$ is close to the support curvature $H^{\mathrm{tr}}_i(\theta_i^*)$.
Thus $H^{\mathrm{tr}}_i(\theta)\approx H^{\mathrm{te}}_i(\theta_i^*)$.

For likelihood-based objectives, the empirical Fisher is commonly used as an approximation to the Hessian in expectation,
motivating $H^{\mathrm{tr}}_i(\theta_i^*)\approx H^{\mathrm{te}}_i(\theta_i^*)\approx F^{(i)}(\theta_i^*)$.
Therefore,
\begin{equation}
H^{\mathrm{tr}}_i(\theta)\,H^{\mathrm{te}}_i(\theta_i^*)\,\delta_i
\;\approx\;
F^{(i)}(\theta_i^*)\,F^{(i)}(\theta_i^*)\,\delta_i
=
\big(F^{(i)}(\theta_i^*)\big)^2\delta_i.
\end{equation}

If $F^{(i)}(\theta_i^*)$ is approximately rank-1, write $F^{(i)}(\theta_i^*)=\rho_i u_i u_i^\top$ with $\rho_i\ge 0$ and $\|u_i\|=1$.
Then
\begin{equation}
\big(F^{(i)}(\theta_i^*)\big)^2
=
(\rho_i u_i u_i^\top)^2
=
\rho_i^2 u_i (u_i^\top u_i) u_i^\top
=
\rho_i^2 u_i u_i^\top
=
\rho_i(\rho_i u_i u_i^\top)
=
\rho_i F^{(i)}(\theta_i^*),
\end{equation}
so $\big(F^{(i)}(\theta_i^*)\big)^2\delta_i \approx \rho_i F^{(i)}(\theta_i^*)\delta_i$.
The scalar $\rho_i$ is the unique nonzero eigenvalue of $F^{(i)}(\theta_i^*)$.
Absorbing $\rho_i$ into the regularizer weight $\lambda$ (and/or the MAML inner-step $\alpha$) yields an EWC-form term.
\section{Additional Experimental details}
\subsection{Task-Incremental Image Generation}
\label{appdx:imagenet}
\subsubsection{Additional Implementation Details}
We adopt a label-conditioned UNet from the \textit{HuggingFace} diffusion library as the denoising model, using default hyperparameters unless otherwise specified.
The architecture consists of four ResNet blocks, with 128 output channels in the first downsampling block and 256 output channels in each subsequent block.
For sampling, we use a DDIM scheduler with 50 inference steps and 1000 diffusion timesteps.
Training hyperparameters are summarized in Table~\ref{tab:imagenetconfig}.
Unless stated otherwise, the same configuration is used across all datasets and tasks.

\begin{table}[h]
\centering
\small
\caption{Training configurations used across methods for ImageNet.}
\label{tab:imagenetconfig}
\begin{tabular}{ll}
\toprule
\textbf{Setting} & \textbf{Value} \\
\midrule
\multicolumn{2}{l}{\textbf{General}} \\
Batch Size & 128 (bilevel: 64 support + 64 query) \\
Optimizer & Adam~\citep{kingma2015adam} (outer); SGD (inner for bilevel) \\
Learning Rate & $2 \times 10^{-4}$ \\
Per-task Gradient Update Steps & 100 epochs/task \\
Seeds Averaged & 3 \\
\midrule
\multicolumn{2}{l}{\textbf{EWC}} \\
Fisher Representation & Rank-1 \cite{wang2025rank1fisher}, estimated over 10{,}000 samples/task \\
EWC $\lambda$ & 15,000 \\
\midrule
\multicolumn{2}{l}{\textbf{Replay}} \\
Buffer Size per Task & 1,300 \\
Replay Ratio & 1.0 (bilevel: 1.0) \\
\midrule
\multicolumn{2}{l}{\textbf{Meta-learning (bilevel) additional hyperparameters}} \\
Adaptation Steps & 1 \\
Inner-loop LR & $1 \times 10^{-4}$ \\
Outer-loop LR & $2 \times 10^{-4}$ \\
\bottomrule
\end{tabular}
\end{table}

\subsubsection{Additional Dataset Details}
We use the downsampled \textbf{ImageNet-1k} dataset (32$\times$32) for computational efficiency while preserving sufficient visual diversity and semantic complexity.
Each sample is normalized to \((-1,1)\) for diffusion training.
We restrict training to the first 500 classes, partitioned into 10 tasks of 50 classes each.
Each task contains approximately 64{,}000 training images.
Task splits:  
$T_1=\{0,\dots,49\},\;
T_2=\{50,\dots,99\},\;
\dots\;
T_{10}=\{450,\dots,499\}$.

\subsection{Continual Robotic Manipulation}
\label{appdx:cw10}

\subsubsection{Diffusion Policy Implementation Details}
We implement a 1D conditional diffusion model over action sequences: given an observation history $o_{1:H}$, the policy generates an action sequence $a_{1:T}$. The denoiser $\epsilon_\theta$ is a 1D conditional U-Net that operates on the action tensor in channel-first form $(B, D_a, T)$ and is FiLM-conditioned on a global context vector formed by concatenating (i) a diffusion-timestep embedding and (ii) the flattened observation history of size $H D_o$. The timestep embedding has dimension 512 and is produced by a sinusoidal positional encoding followed by a 2-layer MLP (512$\rightarrow$2048$\rightarrow$512) with Mish activations. The U-Net uses channel widths $[256,512,1024]$ with kernel size 5 and GroupNorm (8 groups): the encoder has 3 resolution stages, each with two FiLM-conditioned residual Conv1D blocks plus a strided downsampling layer (except the last stage); the bottleneck has 2 additional residual blocks; and the decoder has 2 upsampling stages with skip connections, each with two residual blocks and transposed-convolution upsampling, followed by a final Conv1D block and a $1{\times}1$ projection back to $D_a$. Overall, the noise-prediction network contains 12 FiLM-conditioned residual blocks, and we add multi-head self-attention (8 heads) at U-Net levels 2 and 3 (the 512- and 1024-channel resolutions). For the diffusion process we use a DDIM scheduler with 1000 training timesteps and run 50 denoising steps at inference. Training hyperparameters are summarized in \Cref{tab:cw10config}
\begin{table}[h]
\centering
\small
\caption{Training configurations used across methods for Meta-World CW10.}
\label{tab:cw10config}
\begin{tabular}{ll}
\toprule
\textbf{Setting} & \textbf{Value} \\
\midrule
\multicolumn{2}{l}{\textbf{General}} \\
Batch Size & 256 (bilevel: 128 support + 128 query; effective 256) \\
Optimizer & Adam~\citep{kingma2015adam} (outer); SGD (inner for bilevel) \\
Learning Rate & $2 \times 10^{-4}$ \\
Per-task Update Steps & 50{,}000 (gradient / meta-updates) \\
Seeds Averaged & 4 \\
\midrule
\multicolumn{2}{l}{\textbf{EWC}} \\
Fisher Representation & Rank-1 \cite{wang2025rank1fisher}, estimated over 20{,}000 training samples/task \\
EWC $\lambda$ & 12 \\
\midrule
\multicolumn{2}{l}{\textbf{Replay}} \\
Buffer Size per Task & 2{,}500 rollouts/task \\
Replay Ratio & 1.0 (bilevel: 1.0) \\
\midrule
\multicolumn{2}{l}{\textbf{Meta-learning (bilevel) additional hyperparameters}} \\
Adaptation Steps & 1 \\
Inner-loop LR & $1 \times 10^{-4}$ \\
Outer-loop LR & $2 \times 10^{-4}$ \\
\bottomrule
\end{tabular}
\end{table}

\subsubsection{Additional Task and Dataset Details}
\label{app:cw10-task-details}

We evaluate on a 10-task continual manipulation sequence (CW10) derived from Meta-World.
All tasks share a common low-dimensional state representation and continuous control interface, enabling transfer across tasks.

We use low-dimensional, goal-conditioned state observations $o_t \in \mathbb{R}^{39}$ and continuous actions $a_t \in \mathbb{R}^{4}$.
Following the standard Meta-World state interface, the 4D action consists of a desired end-effector displacement in $\mathbb{R}^3$ and a 1D gripper control.
The 39D observation includes end-effector state, gripper state, object pose information, and goal information (shared across tasks).
For diffusion policy training, we linearly rescale each action dimension to $[-1,1]$ (and clip to the same range at training and sampling time) to keep the action domain bounded and consistent across tasks.

Episodes use a fixed horizon of maximum 200 environment steps.
We report success rate, defined as the fraction of evaluation rollouts that satisfy the task-specific success condition.

\paragraph{Task suite.}
Each task shares the same observation/action interface but requires a distinct manipulation skill:
\begin{itemize}
    \item \textbf{hammer-v1} — The agent uses a hammer to drive a nail into a wall.
    \item \textbf{push-wall-v1} — The agent pushes a block toward a designated target position on a wall.
    \item \textbf{faucet-close-v1} — The agent rotates a faucet handle clockwise to close it.
    \item \textbf{push-back-v1} — The agent pushes an object backward to a specified target location.
    \item \textbf{stick-pull-v1} — The agent grasps a stick tool and uses it to pull an object toward a target region.
    \item \textbf{handle-press-side-v1} — The agent presses a handle on an appliance from the side.
    \item \textbf{push-v1} — The agent pushes an object to a specified target position.
    \item \textbf{shelf-place-v1} — The agent picks up an object and places it onto a shelf.
    \item \textbf{window-close-v1} — The agent closes a window by manipulating the window or its handle.
    \item \textbf{peg-unplug-side-v1} — The agent unplugs a peg from a socket by pulling it out from the side.
\end{itemize}

\subsubsection{Diffusion Policy Data Representation}
\label{app:cw10-config-search}

We conduct a small grid search over three sequence-encoding hyperparameters for the CW10 diffusion policy: sequence length $S$, observation horizon $O$, and action chunk size $A$.
To isolate the effect of these design choices, we use the \emph{same model architecture and training configuration} as in our continual learning experiments, except that we train \emph{one model per task from scratch} (i.e., no task sequence and no continual updates), and then report the average success rate across the 10 CW10 tasks.

As shown in \Cref{tab:cw10-config-grid}, the configuration $(S,O,A)=(8,6,2)$ achieves the highest mean success rate ($0.859$), improving over the next best setting $(8,8,1)$ ($0.808$).
We therefore adopt $S{=}8$, $O{=}6$, and $A{=}2$ as the default CW10 diffusion policy configuration in all main continual learning experiments.

\begin{table}[h]
\centering
\small
\setlength{\tabcolsep}{6pt}
\renewcommand{\arraystretch}{1.15}
\caption{CW10 diffusion policy configuration grid search results.
We report the mean success rate across 10 tasks. Higher is better.
Each configuration is trained from scratch separately for each task, using the same model and training setup as in our continual learning experiments.}
\label{tab:cw10-config-grid}
\begin{tabular}{lcccc}
\toprule
Config & $S$ & $O$ & $A$ & Mean success $\uparrow$ \\
\midrule
\textbf{S8\_O6\_A2}   & 8  & 6  & 2  & \textbf{0.859} \\
S8\_O8\_A1            & 8  & 8  & 1  & 0.808 \\
S8\_O4\_A4            & 8  & 4  & 4  & 0.744 \\
S8\_O4\_A2            & 8  & 4  & 2  & 0.622 \\
S16\_O8\_A8           & 16 & 8  & 8  & 0.511 \\
S16\_O8\_A4           & 16 & 8  & 4  & 0.486 \\
S16\_O12\_A4          & 16 & 12 & 4  & 0.449 \\
S16\_O16\_A1          & 16 & 16 & 1  & 0.326 \\
S32\_O16\_A4          & 32 & 16 & 4  & 0.184 \\
S32\_O24\_A4          & 32 & 24 & 4  & 0.177 \\
S32\_O16\_A8          & 32 & 16 & 8  & 0.121 \\
S32\_O16\_A16         & 32 & 16 & 16 & 0.081 \\
S32\_O24\_A8          & 32 & 24 & 8  & 0.034 \\
\bottomrule
\end{tabular}
\end{table}

\subsection{Compute Efficacy}
\label{appdx:compute}
\begin{table}[t]
\centering
\small
\setlength{\tabcolsep}{6pt}
\renewcommand{\arraystretch}{1.15}
\caption{Model scale and training throughput. Our diffusion backbones are substantially larger than the reduced ResNet used in VR-MCL, and we report the resulting training speed under our first-order implementation versus a second-order (higher-order autodiff) variant.}
\label{tab:modelsize_speed}
\begin{tabular}{lcccccc}
\toprule
Setting & Backbone & Params (M) & \(\times\) vs. 1.09M & Batch size & it/s (FO) & it/s (2nd) \\
\midrule
ImageNet diffusion & UNet & 37.45 & 34.4\(\times\) & 128 & \(\sim\)5 & \(\sim\)1 \\
CW10 diffusion policy & UNet & 83.27 & 76.4\(\times\) & 256 & \(\sim\)5 & \(\sim\)1 \\
VR-MCL from \cite{wu2024meta} & reduced ResNet & 1.09 & 1.0\(\times\) & 32 & -- & -- \\
\bottomrule
\end{tabular}
\end{table}
\begin{table}[htbp]
\centering
\small
\caption{Average training runtime (hours) and GPU used per dataset/method.}
\label{tab:runtime_gpu}
\begin{tabular}{llcc}
\toprule
Evaluation & Method & Hours & GPU \\
\midrule
CW10 & Continual Finetuning & $\sim$ 21 & NVIDIA A40 \\
CW10 & EWC & $\sim$ 29 & NVIDIA A40 \\
CW10 & Replay & $\sim$ 22 & NVIDIA A40 \\
CW10 & Trust Region & $\sim$ 31 & NVIDIA A40 \\
CW10 & FTML & $\sim$ 29 & NVIDIA A40 \\
CW10 & VRMCL & $\sim$ 41 & NVIDIA A40 \\
\midrule
ImageNet & Continual Finetuning & $\sim$ 12 & NVIDIA H200 \\
ImageNet & EWC & $\sim$ 15 & NVIDIA H200 \\
ImageNet & Replay & $\sim$ 22 & NVIDIA H200 \\
ImageNet & Trust Region & $\sim$ 31 & NVIDIA H200 \\
ImageNet & FTML & $\sim$ 29 & NVIDIA H200 \\
ImageNet & VRMCL & $\sim$ 44 & NVIDIA H200 \\

\midrule
\bottomrule
\end{tabular}
\end{table}

Our models are \(34\text{--}76\times\) larger and use substantially larger batches than the original VR-MCL setting \cite{wu2024meta}, making full second-order hypergradients impractical at our target scale (second-order reduces throughput from \(\sim\)5 it/s to \(\sim\)1 it/s, i.e., \(\approx 5\times\) slower) (See \Cref{tab:modelsize_speed}).
Nevertheless, the first-order meta-learning baselines remain scalable and allow us to evaluate variance-reduction and continual-learning behavior under a matched compute budget.
Table~\ref{tab:runtime_gpu} reports the average runtime for each dataset and method (using first order), including evaluation.

\FloatBarrier
\section{Additional Results}
\subsection{Empirical analysis of the local approximations}
\label{app:empirical-analysis}

The MAML-style interpretation in \Cref{sec:cml} relies on locality around old-task optima, agreement between replay and query gradients, and a common Fisher approximation to the support and query Hessians.
We examine these properties on the 83.27M-parameter CW10 diffusion-policy models from our main experiments, then compare the resulting old-task update directly with an exact second-order MAML meta-gradient.

\paragraph{Displacement along protected directions.}
We first examine how the Fisher penalty controls displacement from the task-1 anchor $\theta_1^\star$.
We track $\delta_1=\theta-\theta_1^\star$ at the end of tasks 2 through 10 and compare against a replay-only ablation with the same seed and identical task-1 parameters.
Here, $\theta_i^\star$ denotes the end-of-task parameters of the continual run after training on the mixture of current-task and replay data.
For the rank-1 approximation $F^{(1)}_{\theta_1^\star}\approx\rho_1u_1u_1^\top$, with $\|u_1\|=1$, we measure the protected displacement $|u_1^\top\delta_1|$ and the quadratic trust-region quantity $\delta_1^\top F^{(1)}_{\theta_1^\star}\delta_1$.
The latter corresponds to $c^\star(\mu^\top\delta_1)^2$ in \Cref{eq:trust-region} when $F^{(1)}_{\theta_1^\star}\approx c^\star\mu\mu^\top$.

As shown in \Cref{tab:protected-displacement}, Trust Region reduces the protected displacement by a factor of approximately 7.5 and the quadratic trust-region quantity by a factor of 55 relative to replay alone.
The protected displacement remains stable across all nine subsequent tasks, showing that the penalty maintains locality around the task-1 anchor in the Fisher metric.
The nine per-task protected directions have a mean absolute cosine of 0.11, indicating that the per-task penalties act along largely different directions in parameter space.
Together with the stable protected displacement, this supports the robustness of the Fisher constraint as tasks accumulate.

\begin{table}[htbp]
    \centering
    \small
    \caption{CW10 displacement relative to the task-1 anchor, measured at the ends of tasks 2--10. Trust Region and the replay-only ablation share the same seed and task-1 parameters. Values are rounded.}
    \label{tab:protected-displacement}
    \begin{tabular}{@{}lrr@{}}
        \toprule
        Measurement & Trust Region & Replay only \\
        \midrule
        Protected displacement $|u_1^\top\delta_1|$
            & $4.3\times10^{-3}$ & $3.3\times10^{-2}$ \\
        Trust-region quantity $\delta_1^\top F^{(1)}_{\theta_1^\star}\delta_1$
            & $3.2\times10^{-5}$ & $1.8\times10^{-3}$ \\
        \bottomrule
    \end{tabular}
\end{table}

\paragraph{Gradient and curvature alignment.}
We next examine the gradient and curvature approximations used to connect the two update rules.
Within the replay stream, the replay gradient and the query gradient evaluated at adapted parameters have a cosine similarity of 0.92, supporting the approximation in \Cref{eq:first-conn}.
To evaluate the curvature terms, we compute the top five Hessian eigenpairs at each task optimum on both training and held-out data using Lanczos iterations with exact Hessian-vector products.
The Fisher direction has a cosine similarity of 0.74 with the measured Hessian directions, compared with approximately $10^{-7}$ for the random reference.
Training and held-out Hessians also agree closely, with a top-direction cosine similarity of 0.90 and a leading-eigenvalue ratio of 1.10 (\Cref{tab:local-approximation-probes}).
This agreement in direction and curvature scale supports the common curvature approximation for the support and query terms in \Cref{eq:curvature-identifications}.

\begin{table}[htbp]
    \centering
    \small
    \caption{Gradient and curvature measurements on CW10. Cosine similarities measure directional agreement. The eigenvalue ratio measures agreement in curvature scale, with 1 indicating equal scale. The final row compares the old-task update with an exact second-order one-step MAML meta-gradient using support and query data from the same stream.}
    \label{tab:local-approximation-probes}
    \begin{tabular}{@{}p{0.78\linewidth}r@{}}
        \toprule
        Measurement & Value \\
        \midrule
        Replay gradient vs.\ adapted-query gradient: cosine & 0.92 \\
        Fisher vs.\ Hessian direction: cosine & 0.74 \\
        Training vs.\ held-out Hessian top direction: cosine & 0.90 \\
        Training vs.\ held-out leading Hessian eigenvalue: ratio & 1.10 \\
        Old-task update vs.\ exact MAML meta-gradient: mean cosine & 0.83 \\
        \bottomrule
    \end{tabular}
\end{table}

\paragraph{Alignment with the exact MAML meta-gradient.}
Under the rank-1 Fisher approximation for diffusion models, the curvature reduction $F^2\approx\rho F$ in \Cref{eq:II-to-F2-main} predicts that our old-task update and the MAML meta-gradient share a Fisher-shaped direction.
This is the model setting used in both our image-generation and policy-control experiments.
To test the prediction directly on CW10, we compare the old-task component of our update with an exact second-order one-step MAML meta-gradient at five checkpoints within each task.
Both updates are computed at the same parameters with support and query samples from the same data stream.
The exact MAML calculation agrees with the $(\mathrm{I})-(\mathrm{II})$ expression in \Cref{eq:maml-one-step-two-terms} to within $2\times10^{-7}$.
Our old-task update has a mean cosine similarity of 0.83 with this meta-gradient (\Cref{tab:local-approximation-probes}), providing direct empirical support for the predicted alignment.

\paragraph{Replay quality over longer task sequences.}
\label{app:replay-mismatch}
To examine how the replay-based alignment relates to the original tasks, we repeat the gradient comparison in \Cref{eq:first-conn} with queries drawn from the true old-task distribution.
The cosine similarity is 0.92 initially, reaches 0.55 over the first three tasks, and is 0.28 for tasks 8--10.
These measurements show how replay quality affects agreement with old-task query gradients over longer task sequences.
Together with the displacement results, they illustrate the complementary roles of the two components in the local MAML interpretation.
The Fisher penalty maintains locality along protected directions, while replay supplies the old-task learning signal whose agreement with the original data depends on replay fidelity.
The FMNIST experiments in \Cref{app:fmnist-baseline-sweeps} further examine the role of replay in inner-loop adaptation.

\FloatBarrier

\subsection{Hyperparameter sensitivity and additional baseline experiments}
\label{app:hyperparameter-analysis}

We complement the analysis above with parameter sweeps on CW10 and additional continual meta-learning baseline experiments on Fashion-MNIST.

\subsubsection{CW10 sensitivity analysis}
\label{app:cw10-sensitivity}

We choose the order of magnitude of the EWC coefficient using the Fisher scales $\rho_i=\|F^{(i)}_{\theta_i^\star}\|_{\mathrm{F}}$, following \Cref{sec:cml}.
These scales range from 3 to 20 in measurements taken at already observed task boundaries, so the search uses information available during continual learning.
As in the main experiments, we use a constant coefficient within each benchmark.

At a replay ratio of 1.0, we sweep $\lambda\in\{0,5,12,20\}$, where $\lambda=0$ corresponds to replay alone.
\Cref{tab:cw10-lambda-sweep} shows that the default value $\lambda=12$ achieves the highest success rate and lowest forgetting, at 88.3\% and 4.4, respectively.
This setting improves on replay alone, which achieves a success rate of 85.3\% and forgetting of 8.2.

We next vary the amount of replay to examine whether the benefit of the Fisher penalty persists across replay budgets.
\Cref{tab:cw10-replay-sweep} compares $\lambda=12$ with $\lambda=0$ at four replay ratios from 0.1 to 1.0.
Trust Region improves success rate at every tested ratio, showing that the Fisher penalty complements replay across these settings.

\begin{table}[htbp]
    \centering
    \small
    \caption{CW10 hyperparameter sweeps. Higher success rate (SR, \%) and lower forgetting indicate better performance. The coefficient sweep reports standard errors for $\lambda=0$ and $\lambda=12$.}
    \label{tab:cw10-sensitivity}
    \begin{subtable}[t]{0.48\linewidth}
        \centering
        \caption{EWC coefficient at replay ratio 1.0.}
        \label{tab:cw10-lambda-sweep}
        \setlength{\tabcolsep}{5pt}
        \begin{tabular}{@{}lrr@{}}
            \toprule
            $\lambda$ & SR (\%) & Forgetting \\
            \midrule
            0 & $85.3_{\pm2.0}$ & $8.2_{\pm2.0}$ \\
            5 & 81.0 & 7.3 \\
            12 (default) & $\mathbf{88.3}_{\pm0.4}$ & $\mathbf{4.4}_{\pm0.9}$ \\
            20 & 84.8 & 8.3 \\
            \bottomrule
        \end{tabular}
    \end{subtable}
    \hfill
    \begin{subtable}[t]{0.48\linewidth}
        \centering
        \caption{Success rate (\%) across replay ratios.}
        \label{tab:cw10-replay-sweep}
        \setlength{\tabcolsep}{5pt}
        \begin{tabular}{@{}lrr@{}}
            \toprule
            Replay ratio & $\lambda=0$ & $\lambda=12$ \\
            \midrule
            0.1 & 69.9 & 74.7 \\
            0.5 & 75.3 & 85.2 \\
            0.75 & 82.8 & 85.5 \\
            1.0 & 85.3 & 88.3 \\
            \bottomrule
        \end{tabular}
    \end{subtable}
\end{table}
\FloatBarrier

\paragraph{Fisher estimation sample size.}
Because the penalty is defined by the Fisher direction, we also examine how that direction changes with the number of samples used to estimate it.
Estimates from 2,000 and 20,000 samples have a cosine similarity of 0.96, showing that the penalty direction remains consistent across a tenfold increase in sample size.
This complements the Fisher--Hessian comparison in \Cref{app:empirical-analysis} by assessing the stability of the estimator itself.

\FloatBarrier
\subsubsection{Additional FTML and VR-MCL experiments on Fashion-MNIST}
\label{app:fmnist-baseline-sweeps}

To examine the role of hyperparameter choice and adaptation depth in the continual meta-learning baselines, we run additional FTML and VR-MCL experiments on Fashion-MNIST (FMNIST), where broader sweeps are computationally feasible.
We retain the ImageNet-500 training settings and first-order implementations described in \Cref{sec:experiments}, changing the dataset and the parameters explored in each sweep.
We report final average FID, for which lower values indicate better image quality.

\paragraph{FTML: inner learning rate and adaptation steps.}
We jointly vary the inner learning rate and adaptation depth, comparing one and five inner steps at three learning rates (\Cref{tab:fmnist-ftml-sweep}).
The default learning rate of $10^{-4}$ gives the lowest FID for both adaptation depths.
A single inner step also achieves a lower FID than five steps at all three learning rates.
At the default rate, the corresponding FIDs are 40.8 and 48.0.

\paragraph{VR-MCL: regularization coefficient.}
The VR-MCL sweep varies its regularization coefficient $\alpha$ across four values (\Cref{tab:fmnist-vrmcl-sweep}).
The default coefficient $\alpha=10^{-4}$ gives the lowest FID of 70.2.
Thus, the default hyperparameters used in the ImageNet-500 experiments are also the best settings in the corresponding FMNIST sweeps for both baselines.

\begin{table}[htbp]
    \centering
    \small
    \caption{Final average FID of continual meta-learning baselines on FMNIST. Lower values indicate better image quality. The default hyperparameters are those used in the ImageNet-500 experiments. Bold entries mark the best result for each FTML adaptation depth and for VR-MCL.}
    \label{tab:fmnist-baseline-sweeps}
    \begin{subtable}[t]{0.52\linewidth}
        \centering
        \caption{FTML inner adaptation.}
        \label{tab:fmnist-ftml-sweep}
        \setlength{\tabcolsep}{6pt}
        \begin{tabular}{@{}lrr@{}}
            \toprule
            Inner learning rate & 1 step & 5 steps \\
            \midrule
            $5\times10^{-5}$ & 46.1 & 55.7 \\
            $10^{-4}$ (default) & \textbf{40.8} & \textbf{48.0} \\
            $2\times10^{-4}$ & 57.3 & 70.0 \\
            \bottomrule
        \end{tabular}
    \end{subtable}
    \hfill
    \begin{subtable}[t]{0.43\linewidth}
        \centering
        \caption{VR-MCL regularization.}
        \label{tab:fmnist-vrmcl-sweep}
        \setlength{\tabcolsep}{6pt}
        \begin{tabular}{@{}lr@{}}
            \toprule
            Coefficient $\alpha$ & FID \\
            \midrule
            $10^{-5}$ & 83.4 \\
            $5\times10^{-5}$ & 72.4 \\
            $10^{-4}$ (default) & \textbf{70.2} \\
            $2\times10^{-4}$ & 90.4 \\
            \bottomrule
        \end{tabular}
    \end{subtable}
\end{table}

Trust Region achieves a final average FID of 15.4 on FMNIST, compared with 19.1 for Replay, 40.8 for the best FTML configuration, and 70.2 for the best VR-MCL configuration.
Its advantage therefore persists across the baseline configurations examined here.

The adaptation-depth results also support the replay-based explanation in \Cref{sec:experiments}.
Since each inner step uses replay samples, the higher FID with five steps is consistent with replay errors accumulating during inner optimization.
This complements the CW10 query-gradient comparison in \Cref{app:replay-mismatch} by showing how additional adaptation to replay samples affects generation performance.

\FloatBarrier

\subsection{Permuting Task Order}
\label{app-subsec:permutation}
We conduct an additional experiment by randomly permuting the task order in both CW10 and Imagenet. We see that our methodology is robust to task permutations in \Cref{app-tab:permut_imagenet_fid_forgetting}.

\paragraph{CW10.} Our experiments in \Cref{sec:experiments} were conducted using the standard ordering 
\begin{center}
\small
\texttt{hammer-v1} $\rightarrow$ \texttt{push-wall-v1} $\rightarrow$ \texttt{faucet-close-v1} $\rightarrow$ \texttt{push-back-v1} $\rightarrow$ \texttt{stick-pull-v1} $\rightarrow$ \texttt{handle-press-side-v1} $\rightarrow$ \texttt{push-v1} $\rightarrow$ \texttt{shelf-place-v1} $\rightarrow$ \texttt{window-close-v1} $\rightarrow$ \texttt{peg-unplug-side-v1}
\end{center}
whereas we now rerun the same experiments with a new task ordering
\begin{center}
\small
\texttt{stick-pull-v1} $\rightarrow$ \texttt{push-v1} $\rightarrow$ \texttt{hammer-v1} $\rightarrow$ \texttt{push-back-v1} $\rightarrow$ \texttt{window-close-v1} $\rightarrow$ \texttt{push-wall-v1} $\rightarrow$ \texttt{faucet-close-v1} $\rightarrow$ \texttt{handle-press-side-v1} $\rightarrow$ \texttt{shelf-place-v1} $\rightarrow$ \texttt{peg-unplug-side-v1}
\end{center}

\paragraph{Imagenet-500.} Our experiments in \Cref{sec:experiments} were conducted using the standard ordering 
\begin{center}
\small
$T_1 \to T_2 \to T_3 \to T_4 \to T_5 \to T_6 \to T_7 \to T_8 \to T_9 \to T_{10}$ 
\end{center}
whereas we now rerun the same experiments with a new task ordering
\begin{center}
\small
$T_9 \to T_8 \to T_6 \to T_{10} \to T_3 \to T_4 \to T_7 \to T_2 \to T_5 \to T_1$
\end{center}

\textbf{Note:} These orderings were randomly sampled using \texttt{random.shuffle}.

\begin{table}[!h]
\centering
\scriptsize
\setlength{\tabcolsep}{6pt}
\renewcommand{\arraystretch}{1.2}
\captionsetup{skip=8pt}
\label{app-tab:permut_imagenet_fid_forgetting}
\begin{tabular}{lcccc}
\toprule
\multirow{2}{*}{\textbf{\scriptsize Methods}} &
\multicolumn{2}{c}{\textbf{\scriptsize ImageNet-500}} &
\multicolumn{2}{c}{\textbf{\scriptsize CW10}} \\
\cmidrule(lr){2-3}\cmidrule(lr){4-5}
& \textbf{\scriptsize ${\mathcal{A}\mathrm{FID}}$}$\downarrow$
& \textbf{\scriptsize $\mathcal{F}$}$\downarrow$
& \textbf{\scriptsize ${\mathrm{SR}}$}$\uparrow$
& \textbf{\scriptsize $\mathcal{F}$}$\downarrow$ \\
\midrule
\scriptsize Finetuning   & 91.4  & 55.3  & 26.1\% & 68.3 \\
\scriptsize EWC          & 77.8  & 42.8  & 17.01\% & 70.57 \\
\scriptsize Replay       & 57.5 & 21.5  & 85.8\% &  6.1 \\
\midrule
\scriptsize FTML         & 155.6 & 111.5 & 81.3\% & 6.6 \\
\scriptsize VRMCL        &  146.3 & 104.1  & 74.3\% & 12.8\\
\midrule
\scriptsize Trust Region & \textbf{42.6} & \textbf{8.2} & \textbf{86.8\%} & \textbf{3.4} \\
\bottomrule
\end{tabular}
\caption{Final performance and forgetting on permuted ImageNet-500 and CW10.}
\end{table}
\FloatBarrier

\subsection{Imagenet Per Task Re-Convergence Tables}
We present the re-convergence tables for all Imagenet tasks \Cref{app-tab:recovery2imnet,app-tab:recovery3imnet,app-tab:recovery4imnet,app-tab:recovery5imnet,app-tab:recovery6imnet,app-tab:recovery7imnet,app-tab:recovery8imnet,app-tab:recovery9imnet}.
\begin{table*}[t]
\centering
\tiny
\setlength{\tabcolsep}{3pt}
\begin{tabular}{l *{3}{*{8}{c}}}
\toprule
 & \multicolumn{24}{c}{Imagenet-500} \\
\midrule
Method & \multicolumn{8}{c}{+10\%} & \multicolumn{8}{c}{+20\%} & \multicolumn{8}{c}{+30\%} \\
\cmidrule(lr){2-9}\cmidrule(lr){10-17}\cmidrule(lr){18-25}
 & T3 & T4 & T5 & T6 & T7 & T8 & T9 & T10 & T3 & T4 & T5 & T6 & T7 & T8 & T9 & T10 & T3 & T4 & T5 & T6 & T7 & T8 & T9 & T10 \\
\midrule
Finetune & 76 & -- & -- & -- & -- & -- & -- & -- & 26 & -- & -- & -- & 41 & -- & -- & -- & 16 & -- & -- & -- & 41 & -- & -- & -- \\
EWC & -- & -- & -- & -- & -- & -- & -- & -- & 66 & -- & -- & -- & 9001 & -- & -- & -- & 56 & -- & -- & -- & 7001 & -- & -- & -- \\
Replay & -- & -- & -- & -- & -- & -- & -- & -- & 91 & -- & -- & -- & -- & -- & -- & -- & 71 & 3001 & -- & -- & -- & -- & -- & -- \\
\midrule
FTML & -- & -- & -- & -- & -- & -- & -- & -- & 41 & 36 & -- & -- & -- & -- & -- & -- & 21 & 21 & -- & -- & -- & -- & -- & -- \\
VRMCL & -- & -- & -- & -- & -- & -- & -- & -- & 26 & -- & -- & -- & -- & -- & -- & -- & 26 & -- & -- & -- & -- & -- & -- & -- \\
\midrule
Trust Region & 56 & -- & -- & -- & -- & -- & -- & -- & 16 & 56 & -- & -- & -- & -- & -- & -- & 16 & 51 & 51 & -- & 76 & -- & -- & -- \\
\bottomrule
\end{tabular}
\caption{Steps to re-converge Task~2 under continual learning for Imagenet.}
\label{app-tab:recovery2imnet}
\end{table*}

\begin{table*}[t]
\centering
\tiny
\setlength{\tabcolsep}{3pt}
\begin{tabular}{l *{3}{*{7}{c}}}
\toprule
 & \multicolumn{21}{c}{Imagenet-500} \\
\midrule
Method & \multicolumn{7}{c}{+10\%} & \multicolumn{7}{c}{+20\%} & \multicolumn{7}{c}{+30\%} \\
\cmidrule(lr){2-8}\cmidrule(lr){9-15}\cmidrule(lr){16-22}
 & T4 & T5 & T6 & T7 & T8 & T9 & T10 & T4 & T5 & T6 & T7 & T8 & T9 & T10 & T4 & T5 & T6 & T7 & T8 & T9 & T10 \\
\midrule
Finetune & -- & -- & -- & -- & -- & -- & -- & -- & -- & -- & -- & -- & -- & -- & -- & -- & -- & -- & -- & -- & -- \\
EWC & -- & -- & -- & -- & -- & -- & -- & -- & -- & -- & -- & -- & -- & -- & -- & -- & -- & -- & -- & -- & -- \\
Replay & 701 & -- & -- & -- & -- & -- & -- & 351 & -- & -- & -- & -- & -- & -- & 96 & -- & -- & -- & -- & -- & -- \\
\midrule
FTML & -- & -- & -- & -- & -- & -- & -- & -- & -- & -- & -- & -- & -- & -- & 26 & -- & -- & -- & -- & -- & -- \\
VRMCL & -- & -- & -- & -- & -- & -- & -- & 451 & -- & -- & -- & -- & -- & -- & 71 & -- & -- & -- & -- & -- & -- \\
\midrule
Trust Region & 31 & 51 & 56 & 76 & -- & -- & -- & 31 & 26 & 41 & 46 & -- & -- & -- & 11 & 26 & 36 & 31 & 66 & 41 & -- \\
\bottomrule
\end{tabular}
\caption{Steps to re-converge Task~3 under continual learning for Imagenet.}
\label{app-tab:recovery3imnet}
\end{table*}

\begin{table*}[t]
\centering
\tiny
\setlength{\tabcolsep}{3pt}
\begin{tabular}{l *{3}{*{6}{c}}}
\toprule
 & \multicolumn{18}{c}{Imagenet-500} \\
\midrule
Method & \multicolumn{6}{c}{+10\%} & \multicolumn{6}{c}{+20\%} & \multicolumn{6}{c}{+30\%} \\
\cmidrule(lr){2-7}\cmidrule(lr){8-13}\cmidrule(lr){14-19}
 & T5 & T6 & T7 & T8 & T9 & T10 & T5 & T6 & T7 & T8 & T9 & T10 & T5 & T6 & T7 & T8 & T9 & T10 \\
\midrule
Finetune & -- & -- & -- & -- & -- & -- & -- & -- & -- & -- & -- & -- & -- & -- & -- & -- & -- & -- \\
EWC & -- & -- & -- & -- & -- & -- & -- & -- & -- & -- & -- & -- & -- & -- & -- & -- & -- & -- \\
Replay & -- & -- & -- & -- & -- & -- & -- & -- & -- & -- & -- & -- & -- & -- & -- & -- & -- & -- \\
\midrule
FTML & -- & -- & -- & -- & -- & -- & -- & -- & -- & -- & -- & -- & -- & -- & -- & -- & -- & -- \\
VRMCL & 40001 & -- & -- & -- & -- & -- & 9001 & -- & -- & -- & -- & -- & 6 & 46 & -- & -- & -- & -- \\
\midrule
Trust Region & -- & -- & -- & -- & -- & -- & -- & -- & -- & -- & -- & -- & -- & -- & -- & -- & -- & -- \\
\bottomrule
\end{tabular}
\caption{Steps to re-converge Task~4 under continual learning for Imagenet.}
\label{app-tab:recovery4imnet}
\end{table*}

\begin{table*}[t]
\centering
\tiny
\setlength{\tabcolsep}{3pt}
\begin{tabular}{l *{3}{*{5}{c}}}
\toprule
 & \multicolumn{15}{c}{Imagenet-500} \\
\midrule
Method & \multicolumn{5}{c}{+10\%} & \multicolumn{5}{c}{+20\%} & \multicolumn{5}{c}{+30\%} \\
\cmidrule(lr){2-6}\cmidrule(lr){7-11}\cmidrule(lr){12-16}
 & T6 & T7 & T8 & T9 & T10 & T6 & T7 & T8 & T9 & T10 & T6 & T7 & T8 & T9 & T10 \\
\midrule
Finetune & -- & -- & -- & -- & -- & -- & -- & -- & -- & -- & -- & -- & -- & -- & -- \\
EWC & -- & -- & -- & -- & -- & -- & -- & -- & -- & -- & -- & -- & -- & -- & -- \\
Replay & 46 & -- & -- & -- & -- & 2 & 201 & -- & -- & -- & 2 & 71 & -- & -- & -- \\
\midrule
FTML & -- & -- & -- & -- & -- & -- & -- & -- & -- & -- & 8 & -- & -- & -- & -- \\
VRMCL & -- & -- & -- & -- & -- & -- & -- & -- & -- & -- & -- & -- & -- & -- & -- \\
\midrule
Trust Region & 701 & -- & -- & -- & -- & 51 & 951 & -- & -- & -- & 41 & 551 & -- & -- & -- \\
\bottomrule
\end{tabular}
\caption{Steps to re-converge Task~5 under continual learning for Imagenet.}
\label{app-tab:recovery5imnet}
\end{table*}

\begin{table*}[t]
\centering
\tiny
\begin{subtable}{0.45\textwidth}
\centering
\setlength{\tabcolsep}{3pt}
\begin{tabular}{lcccccc}
\toprule
 & \multicolumn{6}{c}{Imagenet-500} \\
\midrule
Method & \multicolumn{2}{c}{+10\%} & \multicolumn{2}{c}{+20\%} & \multicolumn{2}{c}{+30\%} \\
\cmidrule(lr){2-7}
 & T9 & T10 & T9 & T10 & T9 & T10 \\
\midrule
Finetune & -- & -- & -- & -- & 4 & -- \\
EWC & -- & -- & -- & -- & -- & -- \\
Replay & 851 & -- & 501 & 401 & 151 & 66 \\
\midrule
FTML & -- & -- & -- & -- & 7001 & -- \\
VRMCL & -- & -- & -- & -- & -- & -- \\
\midrule
Trust Region & 2001 & 8001 & 851 & 551 & 41 & 51 \\
\bottomrule
\end{tabular}
\caption{Task 8}
\label{app-tab:recovery8imnet}
\end{subtable}
\hfill
\begin{subtable}{0.45\textwidth}
\centering
\setlength{\tabcolsep}{3pt}
\begin{tabular}{lccccccccc}
\toprule
 & \multicolumn{9}{c}{Imagenet-500} \\
\midrule
Method & \multicolumn{3}{c}{+10\%} & \multicolumn{3}{c}{+20\%} & \multicolumn{3}{c}{+30\%} \\
\cmidrule(lr){2-10}
 & T8 & T9 & T10 & T8 & T9 & T10 & T8 & T9 & T10 \\
\midrule
Finetune & -- & -- & -- & -- & -- & -- & 2 & -- & -- \\
EWC & -- & -- & -- & 56 & -- & -- & 46 & -- & -- \\
Replay & 31 & -- & -- & 26 & 551 & -- & 8 & 151 & 56 \\
\midrule
FTML & -- & -- & -- & 46 & -- & -- & 36 & -- & -- \\
VRMCL & -- & -- & -- & 61 & -- & -- & 46 & -- & -- \\
\midrule
Trust Region & 66 & 2001 & 8001 & 56 & 61 & 51 & 56 & 41 & 26 \\
\bottomrule
\end{tabular}
\caption{Task 7}
\label{app-tab:recovery7imnet}
\end{subtable}

\caption{Steps to re-converge across tasks 7, 8 under continual learning for Imagenet-500.}
\end{table*}
\begin{table*}[t]
\centering
\tiny

\begin{subtable}{0.32\textwidth}
\centering
\setlength{\tabcolsep}{3pt}
\begin{tabular}{lccc}
\toprule
 & \multicolumn{3}{c}{Imagenet-500} \\
\midrule
Method & +10\% & +20\% & +30\% \\
\cmidrule(lr){2-4}
 & T10 & T10 & T10 \\
\midrule
Finetune & 16 & 2 & 2 \\
EWC & 66 & 51 & 16 \\
Replay & 4 & 2 & 2 \\
\midrule
FTML & 11 & 8 & 8 \\
VRMCL & -- & 201 & 81 \\
\midrule
Trust Region & 26 & 21 & 16 \\
\bottomrule
\end{tabular}
\caption{Task 9}
\label{app-tab:recovery9imnet}
\end{subtable}
\hfill
\begin{subtable}{0.6\textwidth}
\centering
\setlength{\tabcolsep}{3pt}
\begin{tabular}{l *{3}{*{4}{c}}}
\toprule
 & \multicolumn{12}{c}{Imagenet-500} \\
\midrule
Method & \multicolumn{4}{c}{+10\%} & \multicolumn{4}{c}{+20\%} & \multicolumn{4}{c}{+30\%} \\
\cmidrule(lr){2-5}\cmidrule(lr){6-9}\cmidrule(lr){10-13}
 & T7 & T8 & T9 & T10 & T7 & T8 & T9 & T10 & T7 & T8 & T9 & T10 \\
\midrule
Finetune & -- & -- & -- & -- & -- & -- & -- & -- & -- & -- & -- & -- \\
EWC & -- & -- & -- & -- & -- & -- & -- & -- & -- & -- & -- & -- \\
Replay & -- & -- & -- & -- & -- & -- & -- & -- & -- & -- & -- & -- \\
\midrule
FTML & -- & -- & -- & -- & -- & -- & -- & -- & -- & -- & -- & -- \\
VRMCL & -- & -- & -- & -- & -- & -- & -- & -- & -- & -- & -- & -- \\
\midrule
Trust Region & -- & -- & -- & -- & -- & -- & -- & -- & -- & -- & -- & -- \\
\bottomrule
\end{tabular}
\caption{Task 6}
\label{app-tab:recovery6imnet}
\end{subtable}

\caption{Steps to re-converge across tasks 6, 9 under continual learning for Imagenet-500.}
\end{table*}


\subsection{CW10 Per Task Re-Convergence Tables}
We present the re-convergence tables for all CW10 tasks \Cref{app-tab:recovery2cw10,app-tab:recovery3cw10,app-tab:recovery4cw10,app-tab:recovery5cw10,app-tab:recovery6cw10,app-tab:recovery7cw10,app-tab:recovery8cw10,app-tab:recovery9cw10}
\begin{table*}[t]
\centering
\tiny
\setlength{\tabcolsep}{3pt}
\begin{tabular}{l *{3}{*{8}{c}}}
\toprule
 & \multicolumn{24}{c}{CW10} \\
\midrule
Method & \multicolumn{8}{c}{99\%} & \multicolumn{8}{c}{90\%} & \multicolumn{8}{c}{80\%} \\
\cmidrule(lr){2-9}\cmidrule(lr){10-17}\cmidrule(lr){18-25}
 & T3 & T4 & T5 & T6 & T7 & T8 & T9 & T10 & T3 & T4 & T5 & T6 & T7 & T8 & T9 & T10 & T3 & T4 & T5 & T6 & T7 & T8 & T9 & T10 \\
\midrule
Finetune & -- & -- & -- & -- & -- & -- & -- & -- & -- & -- & -- & -- & -- & -- & -- & -- & -- & -- & -- & -- & -- & -- & -- & -- \\
EWC & -- & -- & -- & -- & -- & -- & -- & -- & -- & -- & -- & -- & -- & -- & -- & -- & -- & -- & -- & -- & -- & -- & -- & -- \\
Replay & 76 & 46 & 41 & 31 & 5001 & 751 & 9001 & 10001 & 21 & 21 & 21 & 16 & 4 & 46 & 4 & 16 & 10 & 4 & 6 & 4 & 4 & 21 & 4 & 4 \\
\midrule
FTML & 4000 & -- & 550 & -- & -- & -- & -- & -- & 3000 & 250 & 95 & 60 & 4 & -- & 40 & -- & 3000 & 70 & 60 & 55 & 4 & 70 & 4 & 6 \\
VRMCL & 40000 & 30000 & -- & -- & -- & -- & -- & -- & 200 & 60 & 150 & 80 & 4 & 200 & 8 & 35 & 4 & 30 & 150 & 6 & 4 & 75 & 6 & 8 \\
\midrule
Trust Region & 151 & 61 & 46 & 4 & 5001 & 41 & 4 & 6 & 10 & 26 & 16 & 4 & 11 & 21 & 4 & 6 & 8 & 4 & 16 & 4 & 4 & 16 & 4 & 4 \\
\bottomrule
\end{tabular}
\caption{Steps to re-converge Task~2 under continual learning for CW10.}
\label{app-tab:recovery2cw10}
\end{table*}

\begin{table*}[t]
\centering
\tiny
\setlength{\tabcolsep}{3pt}
\begin{tabular}{l *{3}{*{7}{c}}}
\toprule
 & \multicolumn{21}{c}{CW10} \\
\midrule
Method & \multicolumn{7}{c}{99\%} & \multicolumn{7}{c}{80\%} & \multicolumn{7}{c}{70\%} \\
\cmidrule(lr){2-8}\cmidrule(lr){9-15}\cmidrule(lr){16-22}
 & T4 & T5 & T6 & T7 & T8 & T9 & T10 & T4 & T5 & T6 & T7 & T8 & T9 & T10 & T4 & T5 & T6 & T7 & T8 & T9 & T10 \\
\midrule
Finetune & -- & -- & -- & 5000 & -- & 10000 & -- & -- & -- & -- & 4000 & -- & 7000 & -- & 4000 & -- & -- & 3000 & -- & 6000 & -- \\
EWC & -- & -- & -- & -- & -- & -- & -- & -- & -- & -- & -- & -- & -- & -- & -- & -- & -- & -- & -- & -- & -- \\
Replay & 21 & 4 & 4 & 4 & 6 & 4 & 26 & 4 & 4 & 4 & 4 & 4 & 4 & 8 & 4 & 4 & 4 & 4 & 4 & 4 & 6 \\
\midrule
FTML & 35 & 65 & 4 & 4 & 35 & 15 & 35 & 4 & 4 & 4 & 4 & 10 & 4 & 8 & 4 & 4 & 4 & 4 & 4 & 4 & 6 \\
VRMCL & 95 & 400 & 40 & 4 & 40 & 500 & 150 & 20 & 4 & 4 & 4 & 4 & 8 & 45 & 20 & 4 & 4 & 4 & 4 & 8 & 8 \\
\midrule
Trust Region & 16 & 4 & 4 & 4 & 6 & 4 & 16 & 4 & 4 & 4 & 4 & 4 & 4 & 6 & 4 & 4 & 4 & 4 & 4 & 4 & 6 \\
\bottomrule
\end{tabular}
\caption{Steps to re-converge Task~3 under continual learning for CW10.}
\label{app-tab:recovery3cw10}
\end{table*}

\begin{table*}[t]
\centering
\tiny
\setlength{\tabcolsep}{3pt}
\begin{tabular}{l *{3}{*{6}{c}}}
\toprule
 & \multicolumn{18}{c}{CW10} \\
\midrule
Method & \multicolumn{6}{c}{99\%} & \multicolumn{6}{c}{90\%} & \multicolumn{6}{c}{80\%} \\
\cmidrule(lr){2-7}\cmidrule(lr){8-13}\cmidrule(lr){14-19}
 & T5 & T6 & T7 & T8 & T9 & T10 & T5 & T6 & T7 & T8 & T9 & T10 & T5 & T6 & T7 & T8 & T9 & T10 \\
\midrule
Finetune & -- & -- & -- & -- & -- & -- & -- & -- & -- & -- & -- & -- & -- & -- & -- & -- & -- & -- \\
EWC & -- & -- & -- & -- & -- & -- & -- & -- & -- & -- & -- & -- & -- & -- & -- & -- & -- & -- \\
Replay & 36 & 21 & 4 & 8 & 4 & 21 & 10 & 16 & 4 & 8 & 4 & 8 & 8 & 11 & 4 & 4 & 4 & 4 \\
\midrule
FTML & 300 & 55 & 20 & 80 & -- & 600 & 80 & 50 & 20 & 80 & 4 & 20 & 55 & 50 & 10 & 20 & 4 & 20 \\
VRMCL & 55 & 40 & 40 & 50 & 4 & -- & 6 & 40 & 8 & 50 & 4 & -- & 6 & 30 & 8 & 15 & 4 & 75 \\
\midrule
Trust Region & 21 & 16 & 4 & 61 & 4 & 16 & 21 & 16 & 4 & 10 & 4 & 6 & 8 & 10 & 4 & 4 & 4 & 6 \\
\bottomrule
\end{tabular}
\caption{Steps to re-converge Task~4 under continual learning for CW10.}
\label{app-tab:recovery4cw10}
\end{table*}

\begin{table*}[t]
\centering
\tiny
\setlength{\tabcolsep}{3pt}
\begin{tabular}{l *{3}{*{5}{c}}}
\toprule
 & \multicolumn{15}{c}{CW10} \\
\midrule
Method & \multicolumn{5}{c}{99\%} & \multicolumn{5}{c}{90\%} & \multicolumn{5}{c}{80\%} \\
\cmidrule(lr){2-6}\cmidrule(lr){7-11}\cmidrule(lr){12-16}
 & T6 & T7 & T8 & T9 & T10 & T6 & T7 & T8 & T9 & T10 & T6 & T7 & T8 & T9 & T10 \\
\midrule
Finetune & -- & -- & -- & -- & -- & -- & -- & -- & -- & -- & -- & -- & -- & -- & -- \\
EWC & -- & -- & -- & -- & -- & -- & -- & -- & -- & -- & -- & -- & -- & -- & -- \\
Replay & 31 & 6 & 451 & 71 & 76 & 31 & 6 & 31 & 51 & 41 & 26 & 6 & 31 & 4 & 4 \\
\midrule
FTML & -- & -- & -- & -- & -- & 9000 & 50 & -- & -- & -- & 550 & 40 & -- & -- & -- \\
VRMCL & -- & -- & -- & -- & -- & 7000 & 10 & -- & -- & -- & 2000 & 10 & 2000 & -- & 350 \\
\midrule
Trust Region & 96 & 91 & 6001 & 76 & 301 & 86 & 10 & 51 & 16 & 4 & 51 & 6 & 31 & 4 & 4 \\
\bottomrule
\end{tabular}
\caption{Steps to re-converge Task~5 under continual learning for CW10.}
\label{app-tab:recovery5cw10}
\end{table*}
\begin{table*}[t]
\centering
\tiny
\begin{subtable}{0.45\textwidth}
\centering
\setlength{\tabcolsep}{3pt}
\begin{tabular}{l *{3}{*{2}{c}}}
\toprule
 & \multicolumn{6}{c}{CW10} \\
\midrule
Method & \multicolumn{2}{c}{99\%} & \multicolumn{2}{c}{90\%} & \multicolumn{2}{c}{80\%} \\
\cmidrule(lr){2-3}\cmidrule(lr){4-5}\cmidrule(lr){6-7}
 & T9 & T10 & T9 & T10 & T9 & T10 \\
\midrule
Finetune & -- & -- & -- & -- & -- & -- \\
EWC & -- & -- & -- & -- & -- & -- \\
Replay & 36 & 2001 & 4 & 86 & 4 & 4 \\
\midrule
FTML & 65 & -- & 45 & -- & 6 & 500 \\
VRMCL & -- & -- & 8000 & 550 & 4 & 350 \\
\midrule
Trust Region & 96 & -- & 4 & 71 & 4 & 4 \\
\bottomrule
\end{tabular}
\caption{Task 8}
\label{app-tab:recovery8cw10}
\end{subtable}
\hfill
\begin{subtable}{0.45\textwidth}
\centering
\setlength{\tabcolsep}{3pt}
\begin{tabular}{l *{3}{*{3}{c}}}
\toprule
 & \multicolumn{9}{c}{CW10} \\
\midrule
Method & \multicolumn{3}{c}{99\%} & \multicolumn{3}{c}{90\%} & \multicolumn{3}{c}{80\%} \\
\cmidrule(lr){2-4}\cmidrule(lr){5-7}\cmidrule(lr){8-10}
 & T8 & T9 & T10 & T8 & T9 & T10 & T8 & T9 & T10 \\
\midrule
Finetune & -- & -- & -- & -- & -- & -- & -- & -- & -- \\
EWC & -- & -- & -- & -- & -- & -- & -- & -- & -- \\
Replay & 51 & 21 & 36 & 10 & 4 & 4 & 10 & 4 & 4 \\
\midrule
FTML & 550 & 550 & -- & 60 & 6 & 30 & 40 & 4 & 6 \\
VRMCL & 2000 & 75 & -- & 55 & 6 & -- & 30 & 4 & 45 \\
\midrule
Trust Region & 41 & 4 & 36 & 21 & 4 & 4 & 10 & 4 & 4 \\
\bottomrule
\end{tabular}
\caption{Task 7}
\label{app-tab:recovery7cw10}
\end{subtable}

\caption{Steps to re-converge across tasks 7, 8 under continual learning for CW10.}
\end{table*}
\begin{table*}[t]
\centering
\tiny
\begin{subtable}{0.45\textwidth}
\centering
\setlength{\tabcolsep}{3pt}
\begin{tabular}{l *{3}{*{1}{c}}}
\toprule
 & \multicolumn{3}{c}{CW10} \\
\midrule
Method & \multicolumn{1}{c}{99\%} & \multicolumn{1}{c}{90\%} & \multicolumn{1}{c}{80\%} \\
\cmidrule(lr){2-2}\cmidrule(lr){3-3}\cmidrule(lr){4-4}
 & T10 & T10 & T10 \\
\midrule
Finetune & -- & -- & 30 \\
EWC & 30 & 30 & 30 \\
Replay & 4 & 4 & 4 \\
\midrule
FTML & 4 & 4 & 4 \\
VRMCL & 550 & 75 & 10 \\
\midrule
Trust Region & 4 & 4 & 4 \\
\bottomrule
\end{tabular}
\caption{Task 9}
\label{app-tab:recovery9cw10}
\end{subtable}
\hfill
\begin{subtable}{0.45\textwidth}
\centering
\setlength{\tabcolsep}{3pt}
\begin{tabular}{l *{3}{*{4}{c}}}
\toprule
 & \multicolumn{12}{c}{CW10} \\
\midrule
Method & \multicolumn{4}{c}{99\%} & \multicolumn{4}{c}{90\%} & \multicolumn{4}{c}{80\%} \\
\cmidrule(lr){2-5}\cmidrule(lr){6-9}\cmidrule(lr){10-13}
 & T7 & T8 & T9 & T10 & T7 & T8 & T9 & T10 & T7 & T8 & T9 & T10 \\
\midrule
Finetune & 35 & 750 & -- & -- & 4 & 4 & -- & 55 & 4 & 4 & -- & 55 \\
EWC & 40 & -- & -- & -- & 4 & -- & -- & 60 & 4 & 4 & -- & 40 \\
Replay & 4 & 4 & 4 & 6 & 4 & 4 & 4 & 4 & 4 & 4 & 4 & 4 \\
\midrule
FTML & 4 & 4 & 8 & 8 & 4 & 4 & 4 & 6 & 4 & 4 & 4 & 6 \\
VRMCL & 4 & 4 & 4 & 25 & 4 & 4 & 4 & 6 & 4 & 4 & 4 & 4 \\
\midrule
Trust Region & 4 & 4 & 6 & 6 & 4 & 4 & 4 & 4 & 4 & 4 & 4 & 4 \\
\bottomrule
\end{tabular}
\caption{Task 6}
\label{app-tab:recovery6cw10}
\end{subtable}

\caption{Steps to re-converge across tasks 7, 8 under continual learning for CW10.}
\end{table*}
\section{Broader Impacts.}
This work studies continual learning methods that improve retention and rapid re-adaptation in diffusion models and diffusion-based control policies. Positively, such methods could reduce retraining costs, improve sample efficiency, and make adaptive perception and robotics systems more practical in non-stationary settings. However, stronger continual adaptation may also increase misuse risk in generative systems, including more persistent content generation capabilities, and in robotics, where failures under distribution shift could create safety concerns. In addition, results may inherit biases and access restrictions from the underlying datasets and benchmarks, especially ImageNet and robotic simulation environments, so real-world deployment should include dataset auditing, safety constraints, and domain-specific evaluation.
\section{Licenses}
\label{sec:licenses}

We provide details regarding the licenses and access terms of external assets used in this work in Table~\ref{tab:licenses}.

\begin{table}[h!]
\centering
\small
\setlength{\tabcolsep}{6pt}
\renewcommand{\arraystretch}{1.2}
\captionsetup{skip=8pt}
\begin{tabular}{p{0.26\textwidth} p{0.20\textwidth} p{0.44\textwidth}}
\toprule
\textbf{Asset} & \textbf{URL} & \textbf{License} \\
\midrule
\multicolumn{3}{l}{\textbf{Datasets / Benchmarks}} \\
\midrule
ImageNet-500 subset of ImageNet \cite{deng2009imagenet,russakovsky2015imagenet} 
& \href{https://image-net.org/download-images}{Link} 
& ImageNet Terms of Access \\

Meta-World / CW10 source benchmark \cite{yu2020metaworld,wolczyk2021continualworld} 
& \href{https://github.com/Farama-Foundation/Metaworld}{Link} 
& MIT License \\

\midrule
\multicolumn{3}{l}{\textbf{Code / Libraries}} \\
\midrule
Hugging Face \texttt{diffusers}
& \href{https://github.com/huggingface/diffusers}{Link} 
& Apache License 2.0 \\

\bottomrule
\end{tabular}
\caption{Table of licenses.}
\label{tab:licenses}
\end{table}

\clearpage
\section*{NeurIPS Paper Checklist}

\begin{enumerate}

\item {\bf Claims}
    \item[] Question: Do the main claims made in the abstract and introduction accurately reflect the paper's contributions and scope?
    \item[] Answer: \answerYes{} 
    \item[] Justification: See Section 3 and 4.

\item {\bf Limitations}
    \item[] Question: Does the paper discuss the limitations of the work performed by the authors?
    \item[] Answer: \answerYes{} 
    \item[] Justification: See Section 5.

\item {\bf Theory assumptions and proofs}
    \item[] Question: For each theoretical result, does the paper provide the full set of assumptions and a complete (and correct) proof?
    \item[] Answer: \answerYes{} 
    \item[] Justification: See Section 3.

    \item {\bf Experimental result reproducibility}
    \item[] Question: Does the paper fully disclose all the information needed to reproduce the main experimental results of the paper to the extent that it affects the main claims and/or conclusions of the paper (regardless of whether the code and data are provided or not)?
    \item[] Answer: \answerYes{} 
    \item[] Justification: See Section 4.1, 4.2, and Appendix C.

\item {\bf Open access to data and code}
    \item[] Question: Does the paper provide open access to the data and code, with sufficient instructions to faithfully reproduce the main experimental results, as described in supplemental material?
    \item[] Answer: \answerYes{} 
    \item[] Justification: See supplementary materials.

\item {\bf Experimental setting/details}
    \item[] Question: Does the paper specify all the training and test details (e.g., data splits, hyperparameters, how they were chosen, type of optimizer) necessary to understand the results?
    \item[] Answer: \answerYes{} 
    \item[] Justification: See Section 4.1, 4.2, and Appendix C.

\item {\bf Experiment statistical significance}
    \item[] Question: Does the paper report error bars suitably and correctly defined or other appropriate information about the statistical significance of the experiments?
    \item[] Answer: \answerYes{} 
    \item[] Justification: See Section 4.3.

\item {\bf Experiments compute resources}
    \item[] Question: For each experiment, does the paper provide sufficient information on the computer resources (type of compute workers, memory, time of execution) needed to reproduce the experiments?
    \item[] Answer: \answerYes{} 
    \item[] Justification: See Appendix C.
    
\item {\bf Code of ethics}
    \item[] Question: Does the research conducted in the paper conform, in every respect, with the NeurIPS Code of Ethics \url{https://neurips.cc/public/EthicsGuidelines}?
    \item[] Answer: \answerYes{}{} 
    \item[] Justification: We comply with NeurIPS Code of Ethics.

\item {\bf Broader impacts}
    \item[] Question: Does the paper discuss both potential positive societal impacts and negative societal impacts of the work performed?
    \item[] Answer: \answerYes{} 
    \item[] Justification: See Section E.
    
\item {\bf Safeguards}
    \item[] Question: Does the paper describe safeguards that have been put in place for responsible release of data or models that have a high risk for misuse (e.g., pre-trained language models, image generators, or scraped datasets)?
    \item[] Answer: \answerNo{} 
    \item[] Justification: The work proposed in this paper does not pose these specific risks that may warrant safeguards.

\item {\bf Licenses for existing assets}
    \item[] Question: Are the creators or original owners of assets (e.g., code, data, models), used in the paper, properly credited and are the license and terms of use explicitly mentioned and properly respected?
    \item[] Answer: \answerYes{} 
    \item[] Justification: See Table \ref{tab:licenses}.

\item {\bf New assets}
    \item[] Question: Are new assets introduced in the paper well documented and is the documentation provided alongside the assets?
    \item[] Answer: \answerYes{} 
    \item[] Justification: We will release our models to github upon acceptance.

\item {\bf Crowdsourcing and research with human subjects}
    \item[] Question: For crowdsourcing experiments and research with human subjects, does the paper include the full text of instructions given to participants and screenshots, if applicable, as well as details about compensation (if any)? 
    \item[] Answer: \answerNA{} 
    \item[] Justification: This paper does not involve crowdsourcing nor research with human subjects.

\item {\bf Institutional review board (IRB) approvals or equivalent for research with human subjects}
    \item[] Question: Does the paper describe potential risks incurred by study participants, whether such risks were disclosed to the subjects, and whether Institutional Review Board (IRB) approvals (or an equivalent approval/review based on the requirements of your country or institution) were obtained?
    \item[] Answer: \answerNA{} 
    \item[] Justification: This paper does not involve crowdsourcing nor research with human subjects.

\item {\bf Declaration of LLM usage}
    \item[] Question: Does the paper describe the usage of LLMs if it is an important, original, or non-standard component of the core methods in this research? Note that if the LLM is used only for writing, editing, or formatting purposes and does \emph{not} impact the core methodology, scientific rigor, or originality of the research, declaration is not required.
    \item[] Answer: \answerNA{} 
    \item[] Justification: The core work of this paper does not use LLMs, nor are LLMs used as any
important, original, or non-standard components.

\end{enumerate}

\end{document}